\documentclass[journal]{IEEEtran}
\usepackage{cite}

\ifCLASSINFOpdf
  \usepackage[pdftex]{graphicx}
\else
  \usepackage[dvips]{graphicx}
\fi
\usepackage{amsmath}
\ifCLASSOPTIONcompsoc
 \usepackage[caption=false,font=normalsize,labelfont=sf,textfont=sf]{subfig}
\else
 \usepackage[caption=false,font=footnotesize]{subfig}
\fi
\usepackage{url}

\usepackage{amssymb}
\usepackage{threeparttable}
\usepackage[pagebackref=true,breaklinks=true,bookmarks=false,colorlinks]{hyperref}

\begin{document}
%
\title{AttGAN: Facial Attribute Editing by\\Only Changing What You Want}
%
%
%

\author{Zhenliang~He,
        Wangmeng~Zuo,~\IEEEmembership{Senior Member,~IEEE,}
        Meina~Kan,~\IEEEmembership{Member,~IEEE,} \\
        Shiguang~Shan,~\IEEEmembership{Senior Member,~IEEE,}
        and~Xilin~Chen,~\IEEEmembership{Fellow,~IEEE}
}

%
%

\markboth{SUBMITTED MANUSCRIPT TO IEEE TRANSACTIONS ON IMAGE PROCESSING}%
{}
%



\maketitle

\begin{abstract}
Facial attribute editing aims to manipulate single or multiple attributes of a face image, i.e., to generate a new face with desired attributes while preserving other details.
Recently, generative adversarial net (GAN) and encoder-decoder architecture are usually incorporated to handle this task with promising results.
Based on the encoder-decoder architecture, facial attribute editing is achieved by decoding the latent representation of the given face conditioned on the desired attributes.
Some existing methods attempt to establish an attribute-independent latent representation for further attribute editing.
However, such attribute-independent constraint on the latent representation is excessive because it restricts the capacity of the latent representation and may result in information loss, leading to over-smooth and distorted generation.
Instead of imposing constraints on the latent representation, in this work we apply an \textit{attribute classification constraint} to the generated image to just guarantee the correct change of desired attributes, i.e., to ``change what you want''.
Meanwhile, the \textit{reconstruction learning} is introduced to preserve attribute-excluding details, in other words, to ``only change what you want''.
Besides, the \textit{adversarial learning} is employed for visually realistic editing.
These three components cooperate with each other forming an effective framework for high quality facial attribute editing, referred as \textit{AttGAN}.
Furthermore, our method is also directly applicable for \textit{attribute intensity control} and can be naturally extended for \textit{attribute style manipulation}.
Experiments on CelebA dataset show that our method outperforms the state-of-the-arts on realistic attribute editing with facial details~well~preserved.
\end{abstract}


%
\IEEEpeerreviewmaketitle

\section{Introduction}
\begin{figure}[t]
    \centering
    \includegraphics[width=1\linewidth]{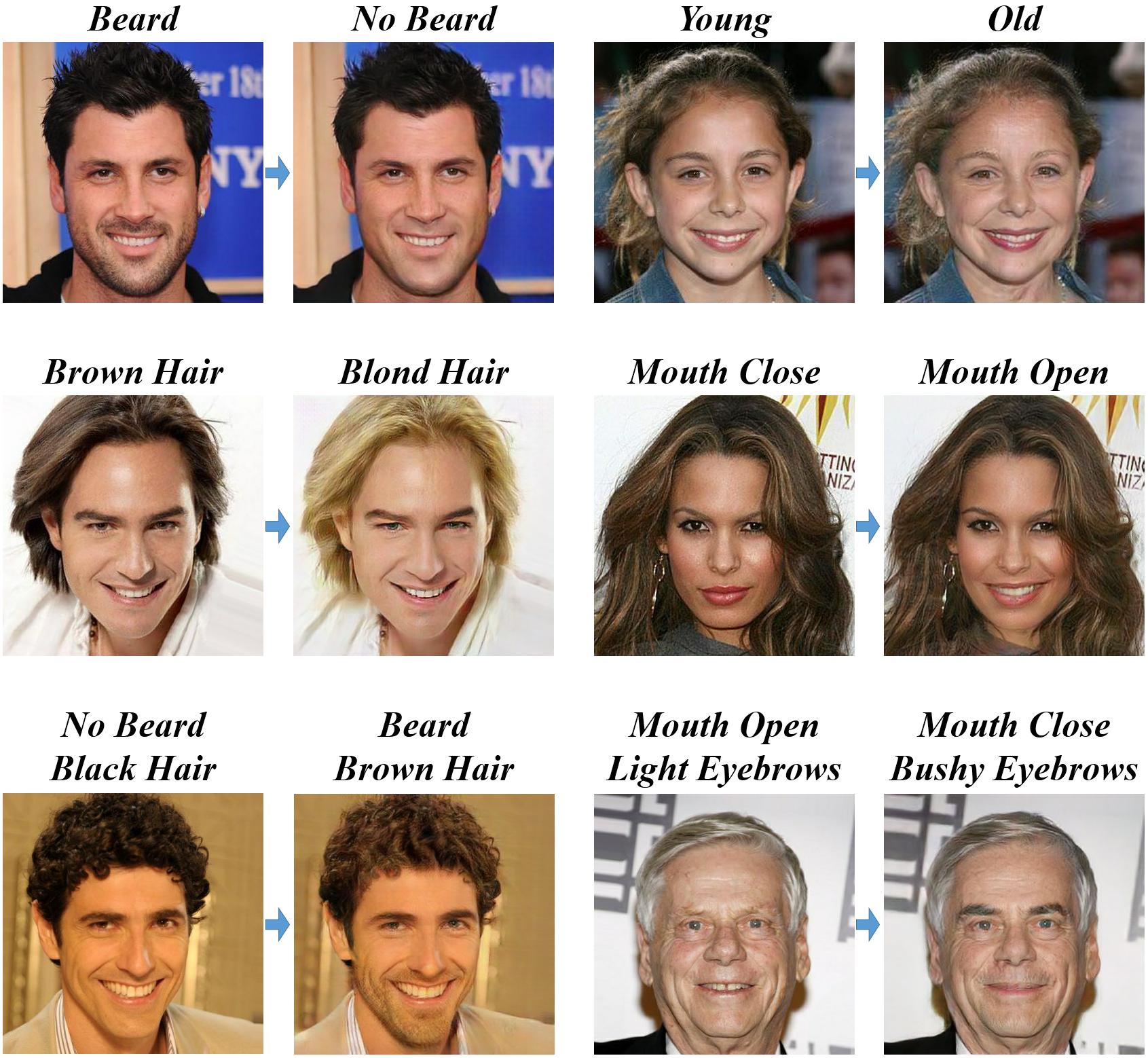}
    \caption{Facial attribute editing results from our AttGAN. Zoom in for better~resolution.}
    \label{fig:first_view}
\end{figure}

%
%
%
%
\IEEEPARstart{T}{his} work investigates the facial attribute editing task, which aims to edit a face image by manipulating single or multiple attributes of interest (e.g., hair color, expression, mustache and age).
For conventional face recognition~\cite{deepid,facenet} and facial attribute prediction~\cite{celeba,ehrlich2016facial} tasks, significant advances have been made along with the development of deep convolutional neural networks (CNNs) and large scale labeled datasets.
However, it is difficult or even impossible to collect labeled images of a same person with varying attributes, thus supervised learning is generally inapplicable for facial attribute editing.
Therefore, researchers turn to generative models such as variational autoencoder (VAE)~\cite{vae} and generative adversarial network (GAN)~\cite{gan}, and make considerable progress on facial attribute editing~\cite{vaegan,icgan,diat,shen2016learning,unit,genegan,fadernetworks,kim2017unsupervised,dnagan,stargan}.

\begin{figure*}[t]
    \centering
    \includegraphics[width=1\linewidth]{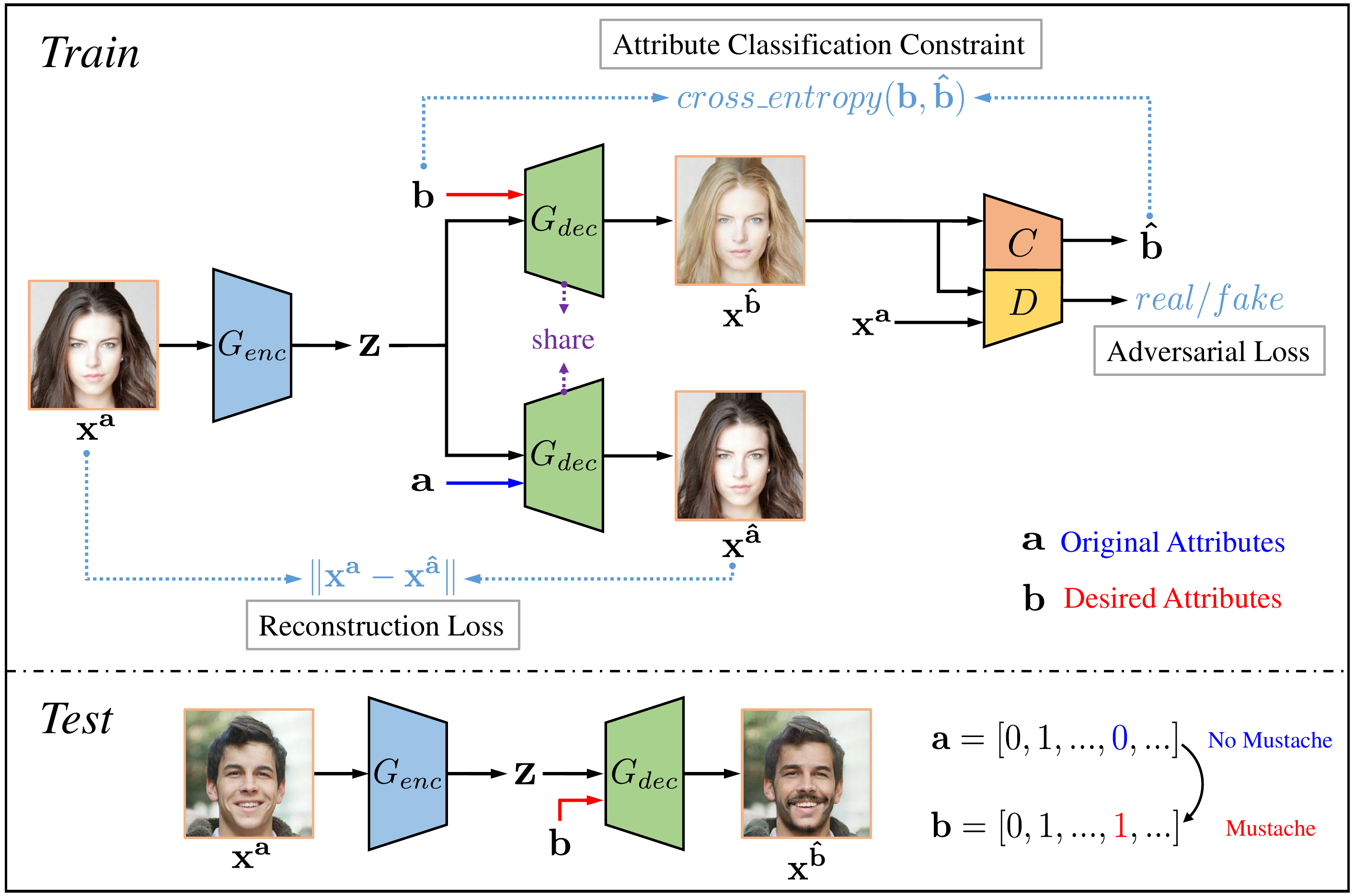}
    \vspace*{-15pt}
    \caption{Overview of our AttGAN, which contains three main components at training: the attribute classification constraint, the reconstruction learning and the adversarial learning. The attribute classification constraint guarantees the correct attribute manipulation on the generated image. The reconstruction learning aims at preserving the attribute-excluding details. The adversarial learning is employed for visually realistic generation.}
    \label{fig:schema}
    \vspace*{-15pt}
\end{figure*}

Some existing methods~\cite{diat,shen2016learning,unit,genegan} use different editing models for different attributes, therefore one has to train numerous models for handling various attribute editing subtasks, which is difficult for real deployment.
For this problem, the encoder-decoder architecture~\cite{vaegan,icgan,fadernetworks,kim2017unsupervised,dnagan} seems to be an effective solution for using a \textit{single} model for \textit{multiple} attribute manipulation.
Therefore, we also focus on the encoder-decoder architecture and develop an effective method for high quality facial attribute editing.

With the encoder-decoder architecture, facial attribute editing is achieved by decoding the latent representation from the encoder conditioned on the expected attributes.
Based on such framework, the key issue of facial attribute editing is \textbf{how to model the relation between the attributes and the face latent representation}.
For this issue, VAE/GAN~\cite{vaegan} represents each attribute as a vector, which is defined as the difference between the mean latent representations of the faces with and without this attribute.
Then, by adding a single or multiple attribute vectors to a face latent representation, the decoded face image from the modified representation is expected to own those attributes.
However, such attribute vector contains highly correlated attributes, thus inevitably leading to unexpected changes of other attributes, e.g., adding blond hair always makes a male become a female because most blond hair objects are female in the training set.
In IcGAN~\cite{icgan}, the latent representation is sampled from a normal distribution independent of the attributes.
In Fader Networks~\cite{fadernetworks}, an adversarial process is introduced to force the latent representation to be invariant to the attributes.
However, the attributes portray the characteristics of a face image, which implies the relation between the attributes and the face latent representation is highly complex and closely dependent.
Therefore, simply imposing the attribute-independent constraint on the latent representation not only restricts its representation ability but also may result in information loss, which is harmful to the~attribute~editing.

With the above limitation analysis of existing methods in mind, we argue that the invariance of the latent representation to the attributes is excessive, and what we need is just the correct editing of attributes.
To this end, instead of imposing the attribute-independence constraint on the latent representation~\cite{icgan,fadernetworks}, we apply an attribute classification constraint to the generated image, just requiring the correct attribute manipulations, i.e., to ``change what you want''.
Therefore in comparison with IcGAN~\cite{icgan} and Fader Networks~\cite{fadernetworks}, the latent representation in our method is constraint free, which guarantees its representation ability and flexibility for further attribute editing.
Besides, we introduce the reconstruction learning for the preservation of the attribute-excluding details\footnote{attribute-excluding details mean the other details of a face image except for the expected attributes, such as face identity, illumination and background.\vspace{-5pt}}, i.e., we aim to ``only change'' the expected attributes while keeping the other details unchanged.
Moreover, the adversarial learning is employed for visually realistic editing.

Our method, referred as AttGAN, can generate visually more pleasing results with fine facial details (see Fig. \ref{fig:first_view}) in comparison with the state-of-the-arts.
Moreover, our AttGAN is directly applicable for attribute intensity control and can be naturally extended for attribute style manipulation.
To sum up, the contribution of this work lies in three folds:
\begin{itemize}
    \item Properly considering the relation between the attributes and the face latent representation under the principle of just satisfying the correct editing objective.
    Our AttGAN removes the strict attribute-independent constraint from the latent representation, and just applies the attribute classification constraint to the generated image to guarantee the correct change of the attributes.
    \item Incorporating the attribute classification constraint, the reconstruction learning and the adversarial learning into a unified framework for high quality facial attribute editing, i.e., the attributes are correctly edited, the attribute-excluding details are well preserved and the whole image is visually realistic.
    \item Promising results of multiple facial attribute editing using a single model. AttGAN outperforms the state-of-the-arts with better perceptual quality for facial attribute editing. Moreover, our method is directly applicable for attribute intensity control and can be naturally extended for attribute style manipulation.
\end{itemize}

\section{Related Work}
\subsection{Facial Attribute Editing}
There are two types of methods for facial attribute editing, the optimization based ones~\cite{cnai,dfi} and the learning based ones~\cite{vaegan,icgan,diat,shen2016learning,unit,genegan,fadernetworks,kim2017unsupervised,stargan}.
Optimization based methods include CNAI~\cite{cnai} and DFI~\cite{dfi}.
To change a given face to the target face with the expected attributes, CNAI~\cite{cnai} defines an attribute loss as the CNN feature difference between the given face and a set of faces with the expected attributes, and then minimizes this loss with respect to the given face.
Based on the assumption that CNN linearizes the manifold of the natural images into an Euclidean feature subspace~\cite{bengio2013better}, DFI~\cite{dfi} first linearly moves the deep feature of the input face along the direction vector between the faces with and without the expected attributes.
Then the facial attribute editing is achieved by optimizing the input face to match its deep feature with the moved feature.
Optimization based methods need to conduct several or even many optimization iterations for each testing image, which are usually time-consuming and unfriendly for real world applications.

Learning based methods are more popular. Li et al.~\cite{diat} present to train a deep identity-aware attribute transfer model to add/remove an attribute to/from a face image by employing an adversarial attribute loss and a deep identity feature loss.
Shen and Liu~\cite{shen2016learning} adopt the dual residual learning strategy to simultaneously train two networks for respectively adding and removing a specific attribute.
GeneGAN~\cite{genegan} swaps a specific attribute between two given images by recombining the information of their latent representation.
These methods~\cite{diat,shen2016learning,unit,genegan}, however, train different models for different attributes (or attribute combinations), leading to large number of models which are also unfriendly for real world applications.

Several learning based methods have been proposed for multiple facial attribute editing with one model.
In VAE/GAN~\cite{vaegan}, GAN~\cite{gan} and VAE~\cite{vae} are combined to learn a latent representation and a decoder.
Then the attribute editing is achieved by modifying the latent representation to own the information of expected attributes and then decoding it.
IcGAN~\cite{icgan} separately trains a cGAN~\cite{cgan} and an encoder, requiring that the latent representation is sampled from a uniform distribution and therefore independent of the attributes.
Then the attribute editing is performed by first encoding an image into the latent representation and then decoding the representation conditioned on the given attributes.
Fader Networks~\cite{fadernetworks} employs an adversarial process on the latent representation of an autoencoder to learn the attribute-invariant representation.
Then, the decoder takes such representation and arbitrary attribute vector as input to generate the edited result.
However, the attribute-independent constraint on the latent representation in IcGAN and Fader Networks is excessive, because it harms the representation ability and may result in information loss, leading to unexpected distortion on the generated images (e.g., over smoothing).
Kim et al.~\cite{kim2017unsupervised} define different blocks of the latent code as the representations of different attributes, and swap several latent code blocks between two given images to achieve multiple attribute swapping.
DNA-GAN~\cite{dnagan} also swap attribute relevant latent blocks between a given pair of images to make ``crossbreed'' images.
Both Kim et al.~\cite{kim2017unsupervised} and DNA-GAN~\cite{dnagan} can be viewed as extensions of GeneGAN~\cite{genegan} for multiple attributes.
StarGAN~\cite{stargan} trains a conditional attribute transfer network via attribute classification loss and cycle consistency loss.
%
StarGAN and our AttGAN are concurrently and independently proposed\footnote{StarGAN first appears  on 2017.11.24 - \url{http://arxiv.org/abs/1711.09020}, and our AttGAN first appears on 2017.11.29 - \url{http://arxiv.org/abs/1711.10678}.} and share some similar objective functions.
Main differences between StarGAN and AttGAN are in two folds: 1) StarGAN uses cycle consistency loss and AttGAN does not include cyclic process or cycle consistency loss, 2) StarGAN trains a conditional attribute transfer network and does not involve any latent representation while AttGAN uses an encoder-decoder architecture and models the relation between the latent representation and the attributes.

Image translation task is closely related to facial attribute editing and some image translation methods are also directly applicable for facial attribute editing.
CycleGAN~\cite{cyclegan} trains two bidirectional transfer models between two image domains by employing the cycle consistency loss and two domain specific adversarial learning processes.
UNIT~\cite{unit} learns to encode the images of two different domains into a common latent space, and then decode the latent representation to the expected domain via the domain specific decoder.
Separating face images with and without the expected attributes into two different domain, one can directly use these methods for facial attribute editing.
However, inability of handling multiple attributes with single model is also the limitation of these domain translation methods.

Our AttGAN is a learning based method for single or multiple facial attribute editing, which is mostly motivated by the encoder-decoder based methods VAE/GAN~\cite{vaegan}, IcGAN~\cite{icgan} and Fader Networks~\cite{fadernetworks}.
We mainly focus on the disadvantages of these three methods on modeling the relation between the latent representation and the attributes, and propose a novel method to solve such problem.

\subsection{Generative Adversarial Networks}
Denote by $p_{data}(\mathbf{x})$ the distribution of the real image $\mathbf{x}$, and $p_{\mathbf{z}}(\mathbf{z})$ the distribution of the input. Generative adversarial net (GAN)~\cite{gan} is a special generative model to learn a generator $G(\mathbf{z})$ to capture the distribution $p_{data}$ via an adversarial process.
Specifically, a discriminator $D$ is introduced to distinguish the generated images from the real ones, while the generator $G(\mathbf{z})$ is updated to confuse the discriminator.
The adversarial process is formulated as a minimax game as
\begin{align}
\resizebox{0.9\hsize}{!}{$
\min\limits_G\max\limits_D ~ \mathbb{E}_{\mathbf{x}\sim p_{data}}[\log D(\mathbf{x})] + \mathbb{E}_{\mathbf{z}\sim p_{\mathbf{z}}}[\log(1 - D(G(\mathbf{z})))].
$}
\end{align}
Theoretically, when the adversarial process reaches the Nash equilibrium, the minimax game attains its global optimum $p_{G(\mathbf{z})} = p_{data}$~\cite{gan}.

GAN is notorious for its unstable training and mode collapse.
DCGAN~\cite{dcgan} uses CNN and batch normalization~\cite{bn} for stable training.
Subsequently, to avoid mode collapse and further enhance the training stability, WGAN~\cite{wgan} minimizes the Wasserstein-1 distance between the generated distribution and the real distribution as
\begin{align} \label{eq:wgan}
\min\limits_G\max\limits_{\|D\|_L\le1} ~ \mathbb{E}_{\mathbf{x} \sim p_{data}}[D(\mathbf{x})] -
\mathbb{E}_{\mathbf{z} \sim p_{\mathbf{z}}}[D(G(\mathbf{z}))],
\end{align}
where $D$ is constrained to be the 1-Lipschitz function implemented by weight clipping.
Furthermore, WGAN-GP~\cite{wgangp} improves WGAN on the implementation of Lipschitz constraint by imposing a gradient penalty on the discriminator instead of weight clipping.
In this work, we adopt WGAN-GP for the adversarial learning.

Several works have been developed for the conditional generation with given attributes or class labels~\cite{cgan,cfgan,acgan,infogan}.
Employing an auxiliary classifier or regressor, both AC-GAN~\cite{acgan} and InfoGAN~\cite{infogan} learn the conditional generation by mapping the generated images back to the conditional signals.
Inspired, in this work, we also map the edited face images back to the given attributes forming the attribute classification constraint. Different from AC-GAN~\cite{acgan}, the generated images do not participate in the training of the auxiliary classifier.

\section{Attribute GAN (AttGAN)} \label{sec:attgan}
This section introduces the AttGAN approach for the editing of binary facial attributes\footnote{each attribute is represented by $1$/$0$ for with/without it and all attributes are represented by a $1$/$0$ sequence.}.
As shown in Fig. \ref{fig:schema}, our AttGAN is comprised of two basic subnetworks, i.e., an encoder $G_{enc}$ and a decoder $G_{dec}$, together with an attribute classifier $C$ and a discriminator $D$.
In the following, we describe the design principles of AttGAN and introduce the objectives for training these components.
Then we present an extension of AttGAN for attribute style manipulation.

\subsection{Testing Formulation} \label{sec:testing_formulation}
Given a face image $\mathbf{x}^\mathbf{a}$ with $n$ binary attributes $\mathbf{a}=[a_1,...,a_n]$, the encoder $G_{enc}$ is used to encode $\mathbf{x}^\mathbf{a}$ into the latent representation, denoted as
\begin{align} \label{eq:z}
\mathbf{z} = G_{enc}(\mathbf{x}^\mathbf{a}).
\end{align}
Then the process of editing the attributes of $\mathbf{x}^\mathbf{a}$ to another attributes $\mathbf{b}=[b_1,...,b_n]$ is achieved by decoding $\mathbf{z}$ conditioned on $\mathbf{b}$, i.e.,
\begin{align}
\mathbf{x}^{\mathbf{\hat{b}}} = G_{dec}(\mathbf{z}, \mathbf{b}),
\end{align}
where $\mathbf{x}^{\mathbf{\hat{b}}}$ is the edited image expected to own the attribute $\mathbf{b}$.
Thus the whole editing process is formulated as
\begin{align} \label{eq:test}
\mathbf{x}^{\mathbf{\hat{b}}} = G_{dec}(G_{enc}(\mathbf{x}^\mathbf{a}), \mathbf{b}).
\end{align}

\subsection{Training Formulation} \label{sec:training_formulation}
It can be seen from Eq.~(\ref{eq:test}) that the attribute editing problem can be formally defined as the learning of the encoder $G_{enc}$ and decoder $G_{dec}$.
This learning problem is unsupervised, because the ground truth of the editing, i.e. $\mathbf{x}^\mathbf{b}$, is unavailable.

On one hand, the editing on the given face image $\mathbf{x}^\mathbf{a}$ is expected to produce a realistic image with attributes $\mathbf{b}$.
For this purpose, an \textbf{attribute classifier} is used to constrain the generated image $\mathbf{x}^{\mathbf{\hat{b}}}$ to correctly own the desired attributes, i.e., the attribute prediction of $\mathbf{x}^{\mathbf{\hat{b}}}$ should be $\mathbf{b}$.
Meanwhile, the \textbf{adversarial learning} is employed on $\mathbf{x}^{\mathbf{\hat{b}}}$ to ensure its visual reality.

On the other hand, an eligible attribute editing should only change those desired attributes, while keeping the other details unchanged.
To this end, the \textbf{reconstruction learning} is introduced to 1) make the latent representation $\mathbf{z}$ conserve enough information for the later recovery of the attribute-excluding details, 2) enable the decoder $G_{dec}$ to restore the attribute-excluding details from $\mathbf{z}$.
Specifically, for the given $\mathbf{x}^{\mathbf{a}}$, the generated image conditioned on its own attributes~$\mathbf{a}$,~i.e.,
\begin{align}
\mathbf{x}^{\mathbf{\hat{a}}}=G_{dec}(\mathbf{z}, \mathbf{a})
\end{align}
should approximate $\mathbf{x}^\mathbf{a}$ itself, i.e., $\mathbf{x}^{\mathbf{\hat{a}}}\rightarrow\mathbf{x}^\mathbf{a}$.

In summary, \textit{the relation between the attributes $\mathbf{a}$/$\mathbf{b}$ and the latent representation $\mathbf{z}$ is implicitly modeled} in two aspects:
1) the interaction between $\mathbf{z}$ and $\mathbf{b}$ in the decoder should produce an realistic image $\mathbf{x}^{\mathbf{\hat{b}}}$ with correct attributes,
and 2) the interaction between $\mathbf{z}$ and $\mathbf{a}$ in the decoder should produce an image $\mathbf{x}^{\mathbf{\hat{a}}}$ approximating the input $\mathbf{x}^\mathbf{a}$ itself.

\begin{figure*}[t]
    \centering
    \subfloat[]{
        \centering
        \includegraphics[width=0.48\linewidth]{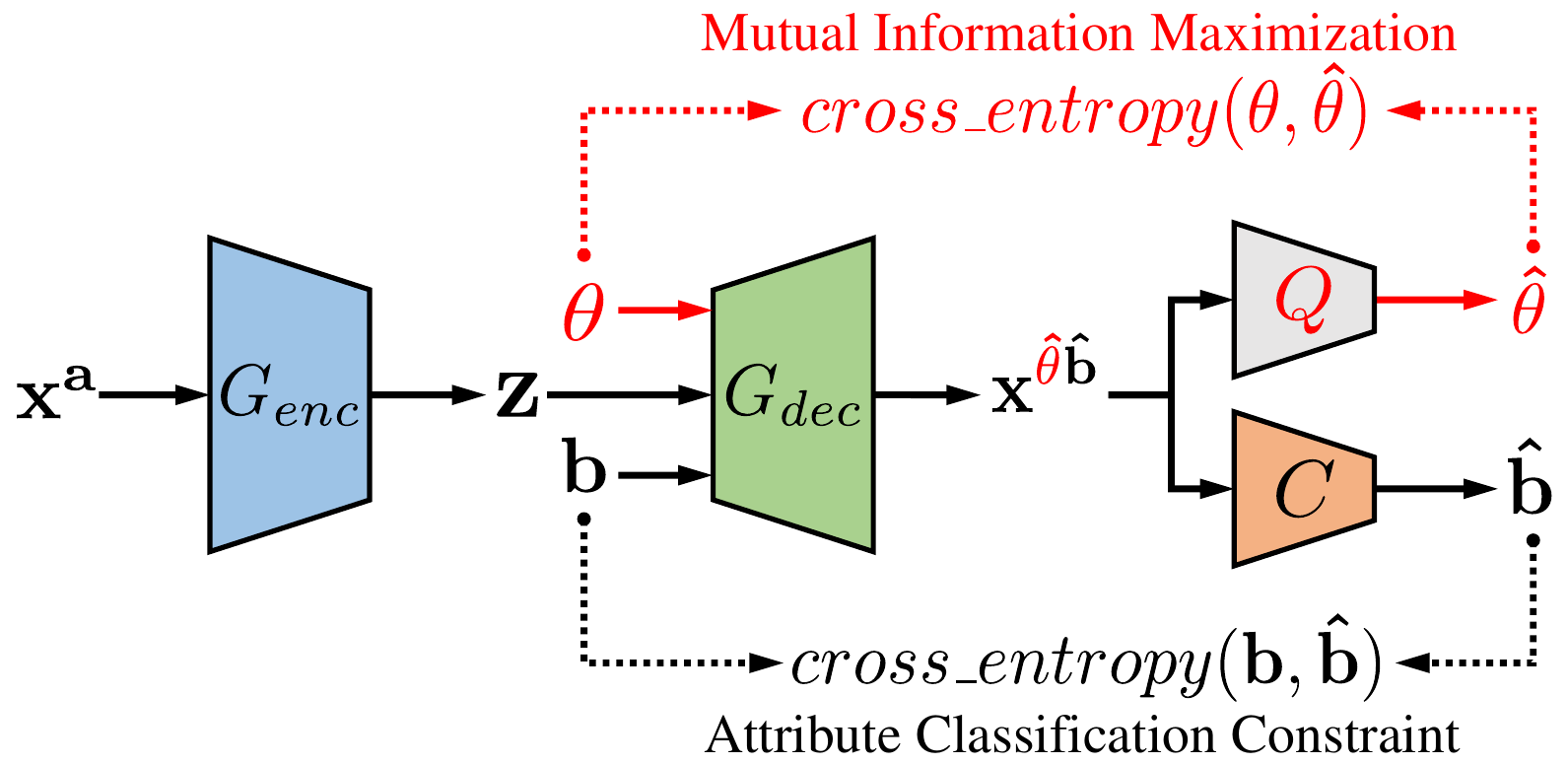}
    }
    \hfill
    \subfloat[]{
        \centering
        \includegraphics[width=0.48\linewidth]{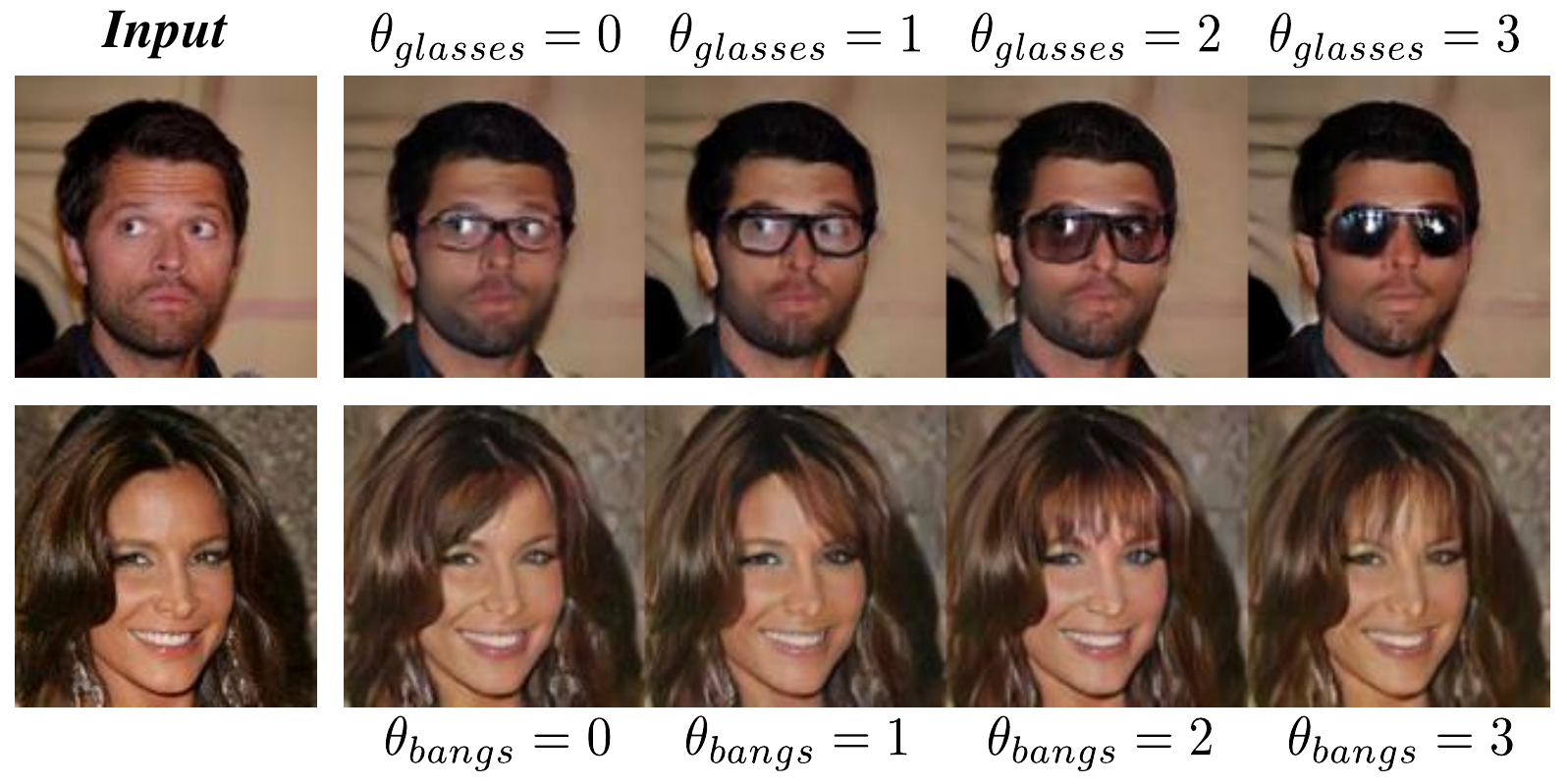}
    }
    \caption{Illustration of AttGAN extension for attribute style manipulation. (a) shows the extended framework based on the original AttGAN. ${\theta}$ denotes the style controllers and $Q$ denotes the style predictor. (b) shows the visual effect of changing attribute style by varying ${\theta}$.}
    \label{fig:cf-attgan}
\end{figure*}

\textbf{Attribute Classification Constraint.}
As mentioned above, it is required that the generated image $\mathbf{x}^{\mathbf{\hat{b}}}$ should correctly own the new attributes $\mathbf{b}$.
%
%
Therefore, we employ an attribute classifier $C$ to constrain the generated image $\mathbf{x}^{\mathbf{\hat{b}}}$ to own the desired attributes, i.e., $C(\mathbf{x}^{\mathbf{\hat{b}}})\rightarrow\mathbf{b}$, formulated as follows,
\begin{align}
\min\limits_{G_{enc},G_{dec}}\mathcal{L}_{cls_g} = \mathbb{E}_{\mathbf{x}^\mathbf{a}\sim p_{data}, \mathbf{b}\sim p_{attr}}[\ell_g(\mathbf{x}^\mathbf{a}, \mathbf{b})],
\end{align}
\begin{align}
\resizebox{0.9\hsize}{!}{$
\ell_g(\mathbf{x}^\mathbf{a}, \mathbf{b})=\sum\limits_{i=1}^n -b_i \log C_i(\mathbf{x}^{\mathbf{\hat{b}}}) - (1 - b_i) \log (1 - C_i(\mathbf{x}^\mathbf{{\hat{b}}})),
$}
\end{align}
where $p_{data}$ and $p_{attr}$ indicate the distribution of real images and the distribution of attributes, $C_i(\mathbf{x}^{\mathbf{\hat{b}}})$ indicates the prediction of the $i^{th}$ attribute, and $\ell_g(\mathbf{x}^\mathbf{a}, \mathbf{b})$ is the summation of binary cross entropy losses of all attributes.

The attribute classifier $C$ is trained on the input images with their original attributes, by the following objective,
\begin{align}
\min\limits_C \mathcal{L}_{cls_c} = \mathbb{E}_{\mathbf{x}^\mathbf{a}\sim p_{data}}[\ell_r(\mathbf{x}^\mathbf{a}, \mathbf{a})],
\end{align}
\begin{align}
\resizebox{0.89\hsize}{!}{$
\ell_r(\mathbf{x}^\mathbf{a}, \mathbf{a})=\sum\limits_{i=1}^n-a_i \log C_i(\mathbf{x}^\mathbf{a}) - (1 - a_i)\log (1 - C_i(\mathbf{x}^\mathbf{a})).
$}
\end{align}

\textbf{Reconstruction Loss.}
Furthermore, the reconstruction learning aims for satisfactory preservation of attribute-excluding details.
To this end, the decoder should learn to reconstruct the input image $\mathbf{x}^\mathbf{a}$ by decoding the latent representation $\mathbf{z}$ conditioned on the original attributes $\mathbf{a}$.
The learning objective is formulated as
\begin{align}
\min\limits_{G_{enc},G_{dec}} \mathcal{L}_{rec} =  \mathbb{E}_{\mathbf{x}^\mathbf{a}\sim p_{data}}[\|\mathbf{x}^\mathbf{a} - \mathbf{x}^{\mathbf{\hat{a}}}\|_1],
\end{align}
where we use the $\ell_1$ loss rather than $\ell_2$ loss to suppress the~blurriness.

\textbf{Adversarial Loss.}
The adversarial learning between the generator (including the encoder and decoder) and discriminator is introduced to make the generated image $\mathbf{x}^{\mathbf{\hat{b}}}$ visually realistic.
Following WGAN~\cite{wgan}, the adversarial losses for the the discriminator and generator are formulated as below,
\newpage
\begin{align}
\resizebox{0.89\hsize}{!}{$
\min\limits_{\|D\|_L\le1} \mathcal{L}_{adv_d}=-\mathbb{E}_{\mathbf{x}^\mathbf{a}\sim p_{data}}D(\mathbf{x}^\mathbf{a})+\mathbb{E}_{\mathbf{x}^\mathbf{a}\sim p_{data},\mathbf{b} \sim p_{attr}}D(\mathbf{x}^{\mathbf{\hat{b}}}),
$}
\end{align}
\begin{align}
\min\limits_{G_{enc},G_{dec}} \mathcal{L}_{adv_g} = -\mathbb{E}_{\mathbf{x}^\mathbf{a}\sim p_{data},\mathbf{b}\sim p_{attr}}[D(\mathbf{x}^{\mathbf{\hat{b}}})],
\end{align}
where $D$ is the discriminator described in Eq.~(\ref{eq:wgan}). The adversarial losses are optimized via WGAN-GP~\cite{wgangp}.


\begin{table*}[t]
    \renewcommand{\arraystretch}{1.0}
    \parbox{\linewidth}{
        \caption{Network Architectures of AttGAN for 128+$^2$ Images.}
        \label{tab:architecture128}
        \begin{threeparttable}
            \centering
            \resizebox{\textwidth}{!}
            {
                \begin{tabular}{|c|c|c|c|}
                    \hline
                    \multicolumn{1}{|c|}{\textbf{Encoder ($G_{enc}$)}} & \multicolumn{1}{|c|}{\textbf{Decoder ($G_{dec}$)}} & \multicolumn{1}{|c|}{\textbf{Discriminator ($D$)}}                      & \multicolumn{1}{|c|}{\textbf{Classifier ($C$)}} \\\hline \hline

                    Conv(64,4,2), BN, Leaky ReLU           & DeConv(1024,4,2), BN, ReLU             & \multicolumn{2}{|c|}{Conv(64,4,2), LN/IN, Leaky ReLU} \\ \hline

                    Conv(128,4,2), BN, Leaky ReLU          & DeConv(512,4,2), BN, ReLU              & \multicolumn{2}{|c|}{Conv(128,4,2), LN/IN, Leaky ReLU} \\ \hline

                    Conv(256,4,2), BN, Leaky ReLU          & DeConv(256,4,2), BN, ReLU              & \multicolumn{2}{|c|}{Conv(256,4,2), LN/IN, Leaky ReLU} \\ \hline

                    Conv(512,4,2), BN, Leaky ReLU          & DeConv(128,4,2), BN, ReLU              & \multicolumn{2}{|c|}{Conv(512,4,2), LN/IN, Leaky ReLU} \\ \hline

                    Conv(1024,4,2), BN, Leaky ReLU         & DeConv(3,4,2), Tanh                    & \multicolumn{2}{|c|}{Conv(1024,4,2), LN/IN, Leaky ReLU} \\ \hline

                                                           &                                        & FC(1024), LN/IN, Leaky ReLU                                       & FC(1024), LN/IN, Leaky ReLU \\ \hline

                                                           &                                        & FC(1)                                                             & FC(13), Sigmoid \\ \hline
                \end{tabular}
            }
        \end{threeparttable}
    }

    \parbox{\linewidth}{
        \caption{Network Architectures of AttGAN for 64$^2$ Images.}
        \label{tab:architecture64}
        \begin{threeparttable}
            \centering
            \resizebox{\textwidth}{!}
            {
                \begin{tabular}{|c|c|c|c|}
                    \hline
                    \multicolumn{1}{|c|}{\textbf{Encoder ($G_{enc}$)}} & \multicolumn{1}{|c|}{\textbf{Decoder ($G_{dec}$)}} & \multicolumn{1}{|c|}{\textbf{Discriminator ($D$)}}                     & \multicolumn{1}{|c|}{\textbf{Classifier ($C$)}} \\ \hline \hline

                    Conv(64,5,2), BN, Leaky ReLU           & DeConv(512,5,2), BN, ReLU              & \multicolumn{2}{|c|}{Conv(64,3,1), LN/IN, Leaky ReLU} \\ \hline

                    Conv(128,5,2), BN, Leaky ReLU          & DeConv(256,5,2), BN, ReLU              & \multicolumn{2}{|c|}{Conv(64,5,2), LN/IN, Leaky ReLU} \\ \hline

                    Conv(256,5,2), BN, Leaky ReLU          & DeConv(128,5,2), BN, ReLU              & \multicolumn{2}{|c|}{Conv(128,5,2), LN/IN, Leaky ReLU} \\ \hline

                    Conv(512,5,2), BN, Leaky ReLU          & DeConv(64,5,2), BN, ReLU               & \multicolumn{2}{|c|}{Conv(256,5,2), LN/IN, Leaky ReLU} \\ \hline

                                                           & DeConv(3,5,1), Tanh                    & \multicolumn{2}{|c|}{Conv(512,5,2), LN/IN, Leaky ReLU} \\ \hline

                                                           &                                        & \multicolumn{2}{|c|}{Conv(512,3,1), LN/IN, Leaky ReLU} \\ \hline

                                                           &                                        & FC(1024), LN/IN, Leaky ReLU                                      & FC(1024), LN/IN, Leaky ReLU \\ \hline

                                                           &                                        & FC(1)                                                            & FC(13), Sigmoid \\ \hline
                \end{tabular}
            }
            \begin{tablenotes}
                \item[*] Conv(d,k,s) and DeConv(d,k,s) denote the convolutional layer and transposed convolutional layer with d as dimension, k as kernel size and s as stride. BN is batch normalization~\cite{bn}, LN is layer normalization~\cite{ln} and IN is instance normalization~\cite{in}.
            \end{tablenotes}
        \end{threeparttable}
    }
\end{table*}

\textbf{Overall Objective.}
By combining the attribute classification constraint, the reconstruction loss and the adversarial loss, an unified attribute GAN (AttGAN) is obtained, which can edit the desired attributes with the attribute-excluding details well preserved.
Overall, the objective for the encoder and decoder is formulated as below,
\begin{align} \label{eq:loss_enc_dec}
\min\limits_{G_{enc},G_{dec}} \mathcal{L}_{enc,dec} = \lambda_1 \mathcal{L}_{rec} +\lambda_2 \mathcal{L}_{cls_g} + \mathcal{L}_{adv_g},
\end{align}
and the objective for the discriminator and the attribute classifier is formulated as below,
\begin{align} \label{eq:loss_dis_cls}
\min \limits_{D,C} \mathcal{L}_{dis,cls} = \lambda_3 \mathcal{L}_{cls_c} + \mathcal{L}_{adv_d},
\end{align}
where the discriminator and the attribute classifier share most layers, $\lambda_1$, $\lambda_2$ and $\lambda_3$ are the hyperparameters for balancing the losses.

\subsection{Why are attribute-excluding details preserved?} \label{sec:discussion}
The above AttGAN design can be viewed as a multi-task leaning of attribute editing task with classification loss and face reconstruction task with reconstruction loss, which share the entire  encoder-decoder network.
However, AttGAN only conducts the reconstruction learning on the generated image conditioned on the original attributes $\mathbf{a}$, why the preservation ability of attribute-excluding details can be generalized to the generation conditioned on another attributes $\mathbf{b}$?
We suggest the reason is that, AttGAN transfers the detail preservation ability from the face reconstruction task to the attribute editing task.
Since these two tasks share the same input domain and output domain, they are very similar tasks with tiny \textit{transferability gap}~\cite{yosinski2014transferable} between them.
Therefore, the detail preservation ability learned from the face reconstruction task can be easily transfered to the attribute editing task.
Besides, these two tasks are learned simultaneously, therefore such transfer is dynamic and the attribute editing learning does not flush the ability of facial detail reconstruction.


\subsection{Extension for Attribute Style Manipulation}
In Sec. \ref{sec:attgan}, the attributes are binary represented, i.e., ``with'' or ``without'', which is stiff for real world applications.
However, for example, in most cases what one is interested in is adding a certain style of eyeglasses such as sunglasses or thin rim glasses, rather than just with/without eyeglasses.
This problem is more difficult because the labeled data with attribute style is unavailable.
To enable our AttGAN to manipulate the attribute style, a set of style controllers $\theta=[\theta_1, \cdots, \theta_i, \cdots, \theta_n]$ is introduced.
Then following \cite{infogan} and \cite{cfgan}, we bind each $\theta_i$ and the $i^{th}$ attribute, and maximize the mutual information between the controllers and the output images to make them highly correlated.
As a result, such high correlation enables each $\theta_i$ to control the corresponding attribute of the output images.

As shown in Fig.~\ref{fig:cf-attgan}, based on the original AttGAN, we add style controllers ${\theta}$ and a style predictor $Q$, and the attribute editing is reformulated as
\begin{align}
\mathbf{x}^{{\hat{\theta}}\mathbf{\hat{b}}}=G_{dec}(G_{enc}(\mathbf{x}^\mathbf{a}),{\theta},\mathbf{b}),
\end{align}
where $\mathbf{x}^{{\hat{\theta}}\mathbf{\hat{b}}}$ is expected to not only own the attribute $\mathbf{b}$, but also be in the style specified by ${\theta}$.
According to~\cite{infogan}, the mutual information between ${\theta}$ and the output images $\mathbf{x}^*$\footnote{$\mathbf{x}^*\sim G_{dec}(G_{enc}(\mathbf{x}^\mathbf{a}),{\theta},\mathbf{b}),\mathbf{x}^\mathbf{a}\sim p_{data},\mathbf{b}\sim p_{attr},{\theta_i}\sim p_{\theta_i}=\mathrm{Cat}(n_i, \frac{1}{n_i})$, where $n_i$ is predefined number of styles for the $i^{th}$ attribute.} is~obtained~by
\begin{align}
\resizebox{0.89\hsize}{!}{$
I({\theta};\mathbf{x}^*)=\max\limits_Q\mathbb{E}_{\theta\sim p(\theta),\mathbf{x}^*\sim p(\mathbf{x}^*|{\theta})}[\log Q({\theta}|\mathbf{x}^*)]+const.,
$}
\end{align}
and is maximized as
\begin{align}
\max\limits_{G_{enc},G_{dec}}I({\theta};\mathbf{x}^*),
\end{align}
where we achieve the mutual information maximization by optimizing the encoder $G_{enc}$ and decoder $G_{dec}$.
By correlating the output images with the style controllers via mutual information maximization, AttGAN is able to manipulate the attributes in different styles in a totally unsupervised way.

\section{Implementation Details}
Our AttGAN is implemented by the machine learning system Tensorflow~\cite{abadi2016tensorflow} and the code is publicly available at \url{https://github.com/LynnHo/AttGAN-Tensorflow}. Please refer to the website for more implementation details.

\textbf{Network Architecture.}
Table~\ref{tab:architecture128} and Table~\ref{tab:architecture64} shows the detailed network architectures of our AttGAN.
The discriminator $D$ is a stack of convolutional layers followed by fully connected layers, and the classifier $C$ has a similar architecture and shares all convolutional layers with $D$.
The encoder $G_{enc}$ is a stack of convolutional layers and the decoder $G_{dec}$ is a stack of transposed convolutional layers.
We also employ the U-Net~\cite{unet} like symmetric skip connections between the encoder and decoder, which have been shown to produce high quality results on the image translation task~\cite{pix2pix}.
%
Architectures for $64\times 64$ images are used in the comparisons with VAE/GAN~\cite{vaegan} and IcGAN~\cite{icgan}, and architectures for $128\times 128$ images are used in the comparisons with StarGAN~\cite{stargan}, Fader Networks~\cite{fadernetworks}, Shen et al.~\cite{shen2016learning} and CycleGAN~\cite{cyclegan}.
$384\times 384$ images are shown in other experiments for better~visual~effect.

\begin{figure*}[p]
    \parbox[c][0.99\textheight]{1\linewidth}{
        \subfloat[Comparisons with VAE/GAN~\cite{vaegan} and IcGAN~\cite{icgan} on editing (inverting) specified attributes.]{
            \centering
            \includegraphics[width=1\linewidth]{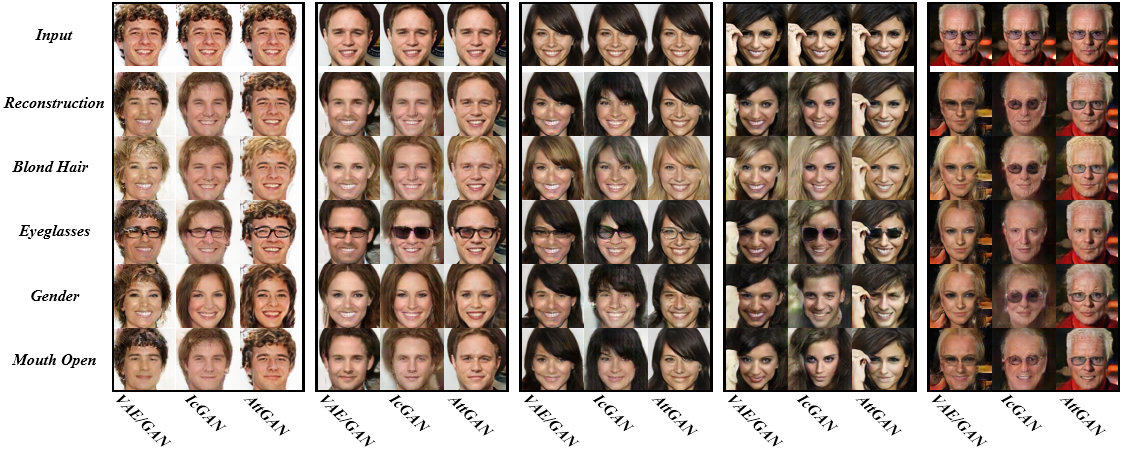}
            \label{fig:compare}
        }
        \vfill
        \subfloat[Comparisons with StarGAN~\cite{stargan} on editing (inverting) specified attributes. Zoom in for better resolution.]{
            \centering
            \includegraphics[width=1\linewidth]{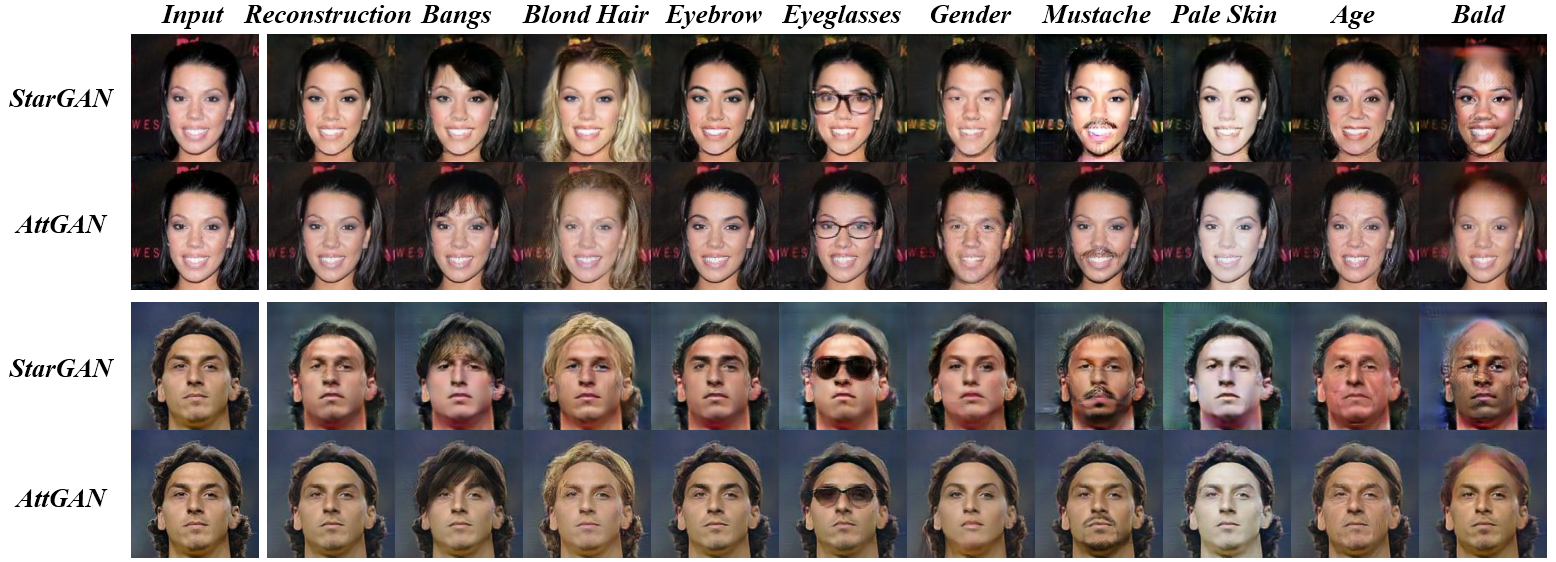}
            \label{fig:compare_stargan}
        }
        \vfill
        \subfloat[Comparisons with Fader Networks~\cite{fadernetworks}, Shen et al.~\cite{shen2016learning} and CycleGAN~\cite{cyclegan} on editing (inverting) specified attributes. Zoom in for better resolution.]{
            \centering
            \includegraphics[width=1\linewidth]{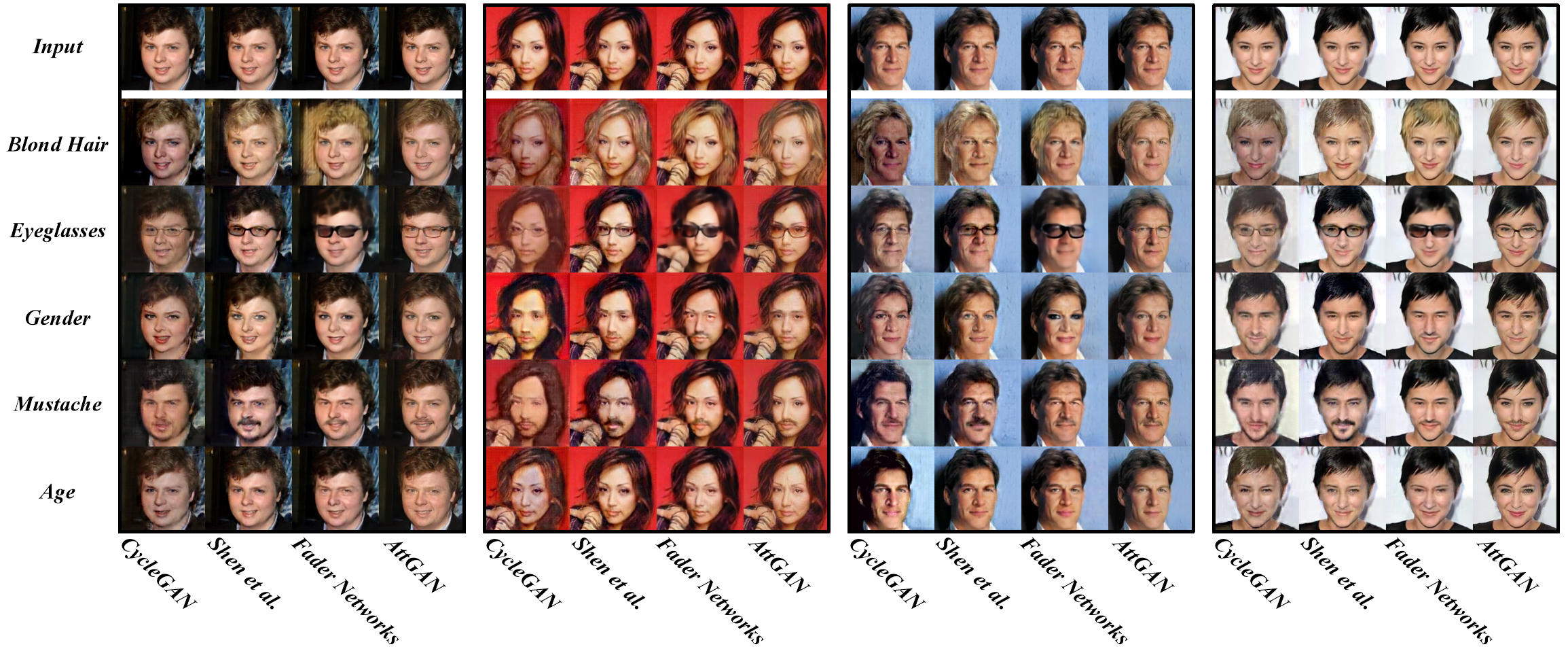}
            \label{fig:compare_3}
        }
        \caption{Results of single facial attribute editing. For each specified attribute, the facial attribute editing here is to \textbf{invert} it, e.g., to edit female to male, male to female, mouth open to mouth close, and mouth close to mouth open etc.}
    }
\end{figure*}

\begin{figure*}[t]
    \centering
    \includegraphics[width=1\linewidth]{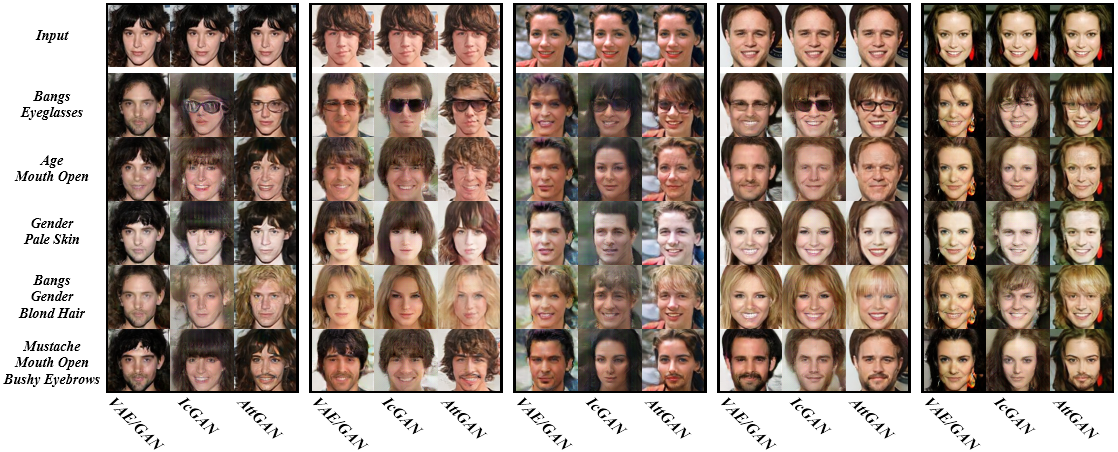}
    \caption{Comparisons of multiple facial attribute editing among our AttGAN, VAE/GAN~\cite{vaegan} and IcGAN~\cite{icgan}. For each specified attribute combination, the facial attribute editing here is to \textbf{invert} each attribute in that combination.}
    \label{fig:compare_multi}
\end{figure*}

\begin{figure*}[t]
    \centering
    \includegraphics[width=1\linewidth]{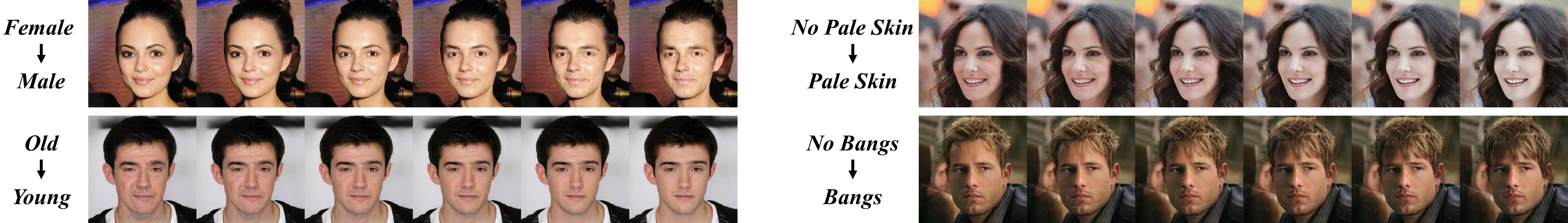}
    \caption{Illustration of attribute intensity control. Zoom in for better resolution.}
    \label{fig:slide}
\end{figure*}

\begin{figure*}[t]
    \centering
    \includegraphics[width=0.95\linewidth]{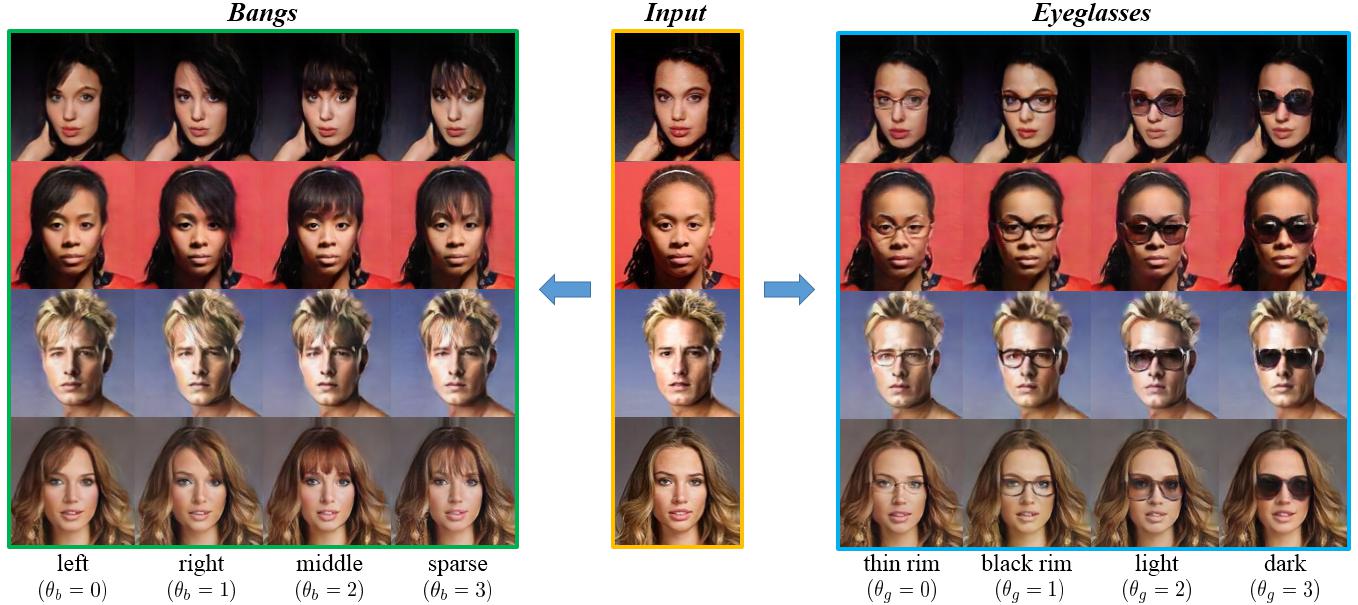}
    \vspace{-10pt}
    \caption{Exemplar results of attribute style manipulation by using our extended AttGAN.}
    \label{fig:style}
\end{figure*}

\begin{figure*}[t]
    \centering
    \subfloat[Attribute Editing Accuracy (Higher the Better)]{
        \centering
        \includegraphics[width=0.47\linewidth]{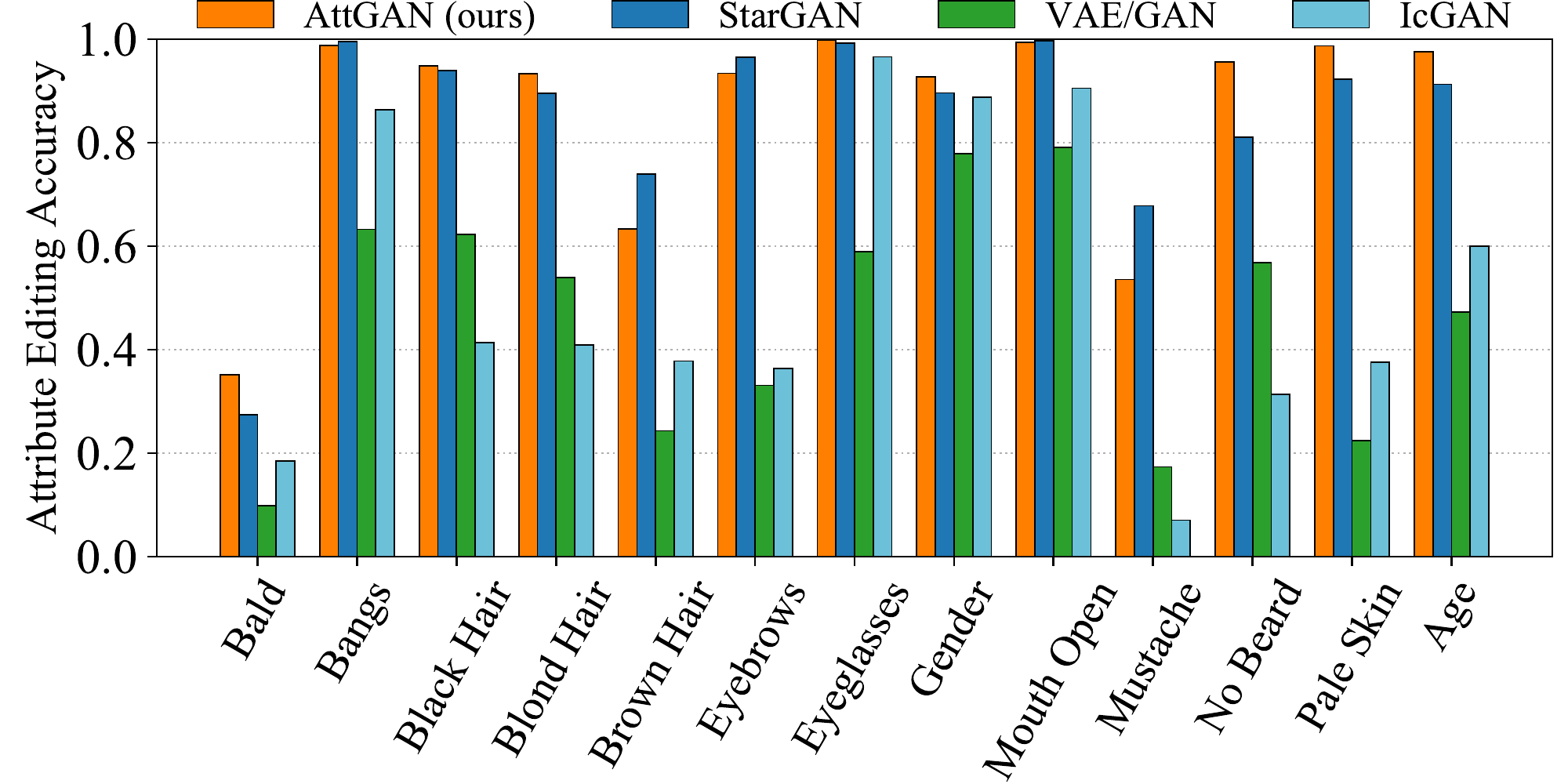}
        \label{fig:attribute_accuracy_1}
    }
    \hfill
    \subfloat[Attribute Preservation Error (Lower the Better)]{
        \centering
        \includegraphics[width=0.47\linewidth]{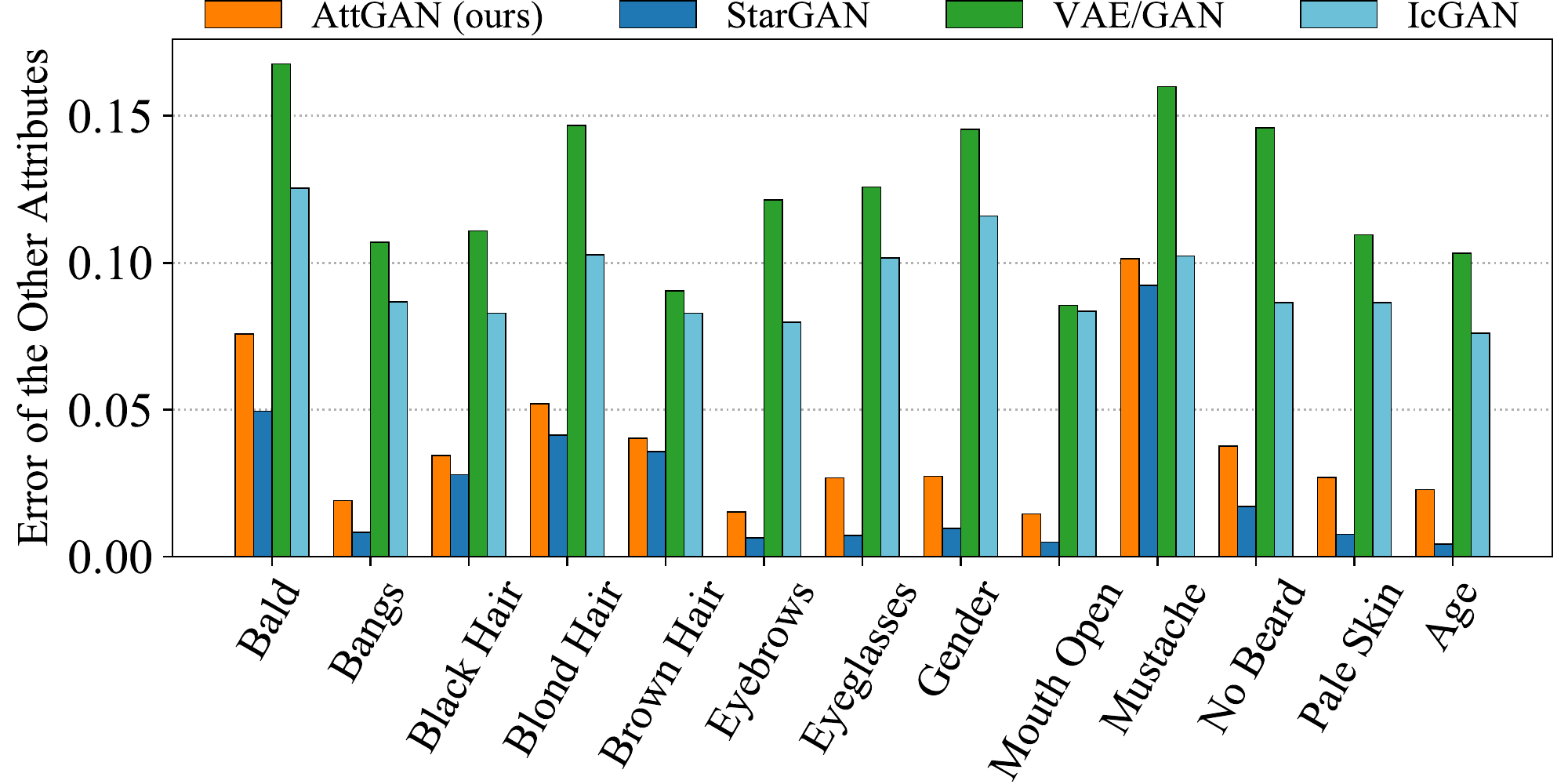}
        \label{fig:attribute_error_1}
    }
    \vspace{-4pt}
    \caption{Comparisons among StarGAN~\cite{stargan}, VAE/GAN~\cite{vaegan}, IcGAN~\cite{icgan} and our AttGAN in terms of (a) facial attribute editing accuracy and (b) preservation error of the other attributes.}

    \centering
    \subfloat[Attribute Editing Accuracy (Higher the Better)]{
        \centering
        \includegraphics[width=0.47\linewidth]{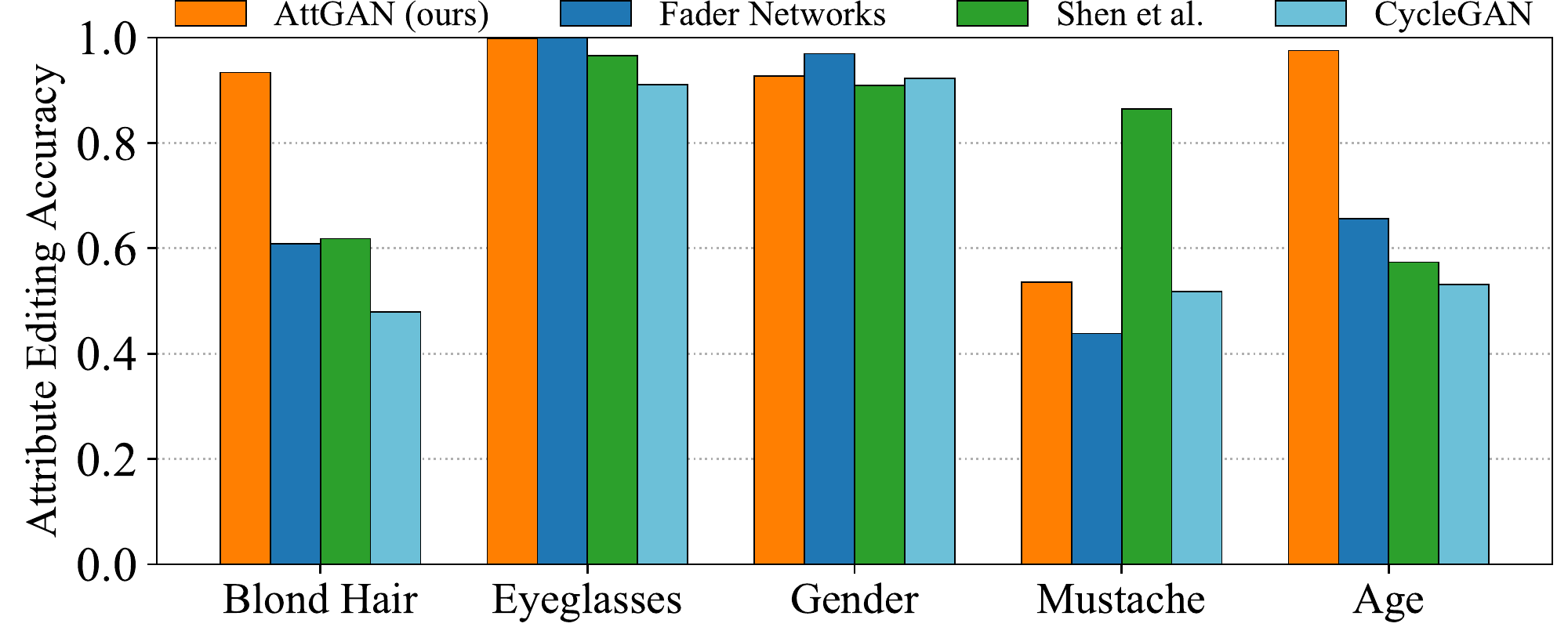}
        \label{fig:attribute_accuracy_2}
    }
    \hfill
    \subfloat[Attribute Preservation Error (Lower the Better)]{
        \centering
        \includegraphics[width=0.47\linewidth]{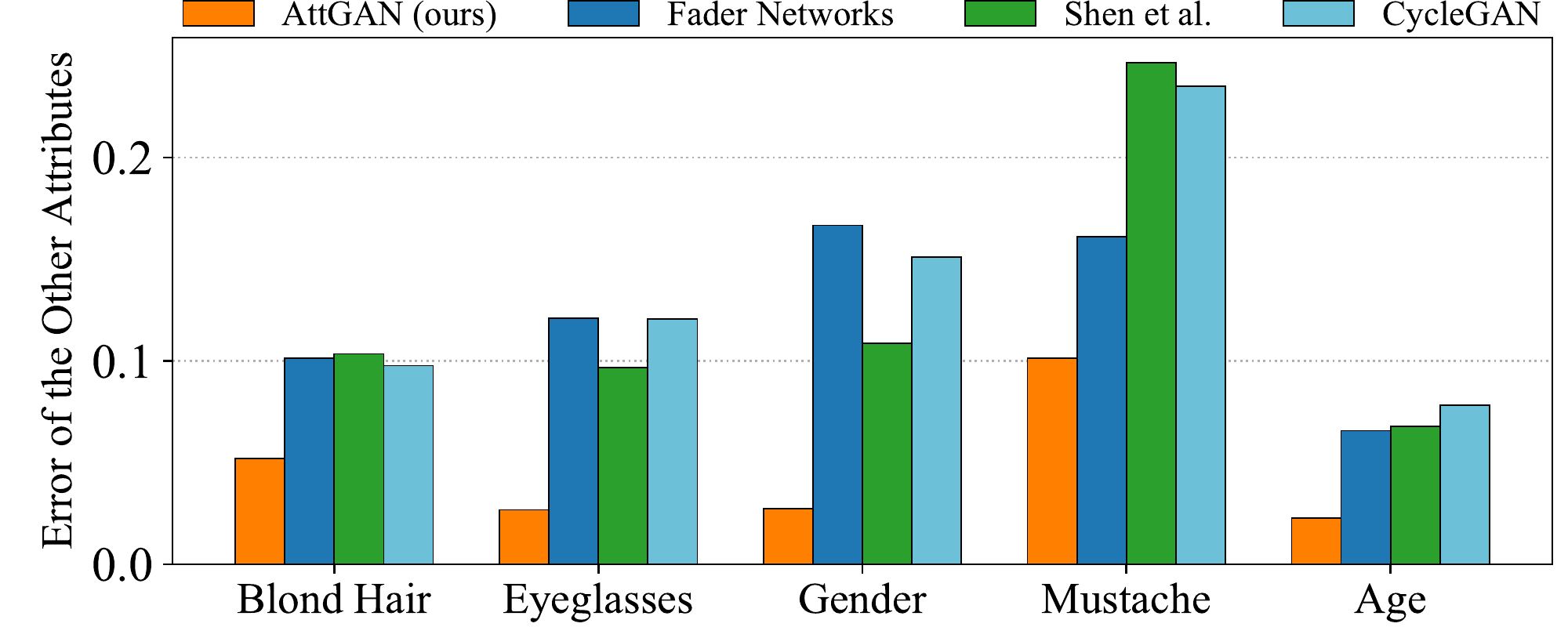}
        \label{fig:attribute_error_2}
    }
    \vspace{-4pt}
    \caption{Comparisons among Fader Networks~\cite{fadernetworks}, Shen et al.~\cite{shen2016learning}, CycleGAN~\cite{cyclegan} and our AttGAN in terms of (a) facial attribute editing accuracy and (b) preservation error of the other attributes.}
    \vspace{-11pt}
\end{figure*}

\textbf{Training Details.}
The model is trained by Adam optimizer~\cite{adam} ($\beta_1=0.5, \beta_2=0.999$) with the batch size of 32 and the learning rate of 0.0002.
The coefficients for the losses in Eq. (\ref{eq:loss_enc_dec}) and Eq. (\ref{eq:loss_dis_cls}) are set as: $\lambda_1=100$, $\lambda_2=10$, and $\lambda_3=1$, which aims to make the
loss values be in the same order of magnitude.

\section{Experiments}
\textbf{Dataset.}
We evaluate the proposed AttGAN on CelebA~\cite{celeba} dataset, which contains two hundred thousand images, each of which has annotation of 40 binary attributes (with/without).
Thirteen attributes with strong visual impact are chosen in all our experiments, including ``Bald'', ``Bangs'', ``Black Hair'', ``Blond Hair'', ``Brown Hair'', ``Bushy Eyebrows'', ``Eyeglasses'', ``Gender'', ``Mouth Open'', ``Mustache'', ``No Beard'', ``Pale Skin'' and ``Age'', which cover most attributes used in the existing works.
Officially, CelebA is separated into training set, validation set and testing set. We use the training set and validation set together to train our model while using the testing set for evaluation.

\textbf{Methods.}
Under the same experimental settings, we compare our AttGAN with two closely related works: VAE/GAN~\cite{vaegan} and IcGAN~\cite{icgan}.
We also compare AttGAN with the concurrent work StarGAN~\cite{stargan}.
All of VAE/GAN, IcGAN, StarGAN and our AttGAN are trained to \textit{handle thirteen attributes with a single model}.
Besides, we compare our AttGAN with the recent Fader Networks~\cite{fadernetworks} (also closely related), Shen et al.~\cite{shen2016learning} and CycleGAN~\cite{cyclegan}.
Shen et al. and CycleGAN can handle only one attribute with one model.
Although Fader Networks is capable for multiple attribute editing with one model, in practice, multiple attribute setting makes the results blurry.
Therefore, for these three baselines, \textit{each attribute has its own specific model}.
VAE/GAN\footnote{VAE/GAN: \url{https://github.com/andersbll/autoencoding\_beyond\_pixels}}, IcGAN\footnote{IcGAN: \url{https://github.com/Guim3/IcGAN}}, StarGAN\footnote{StarGAN: \url{https://github.com/yunjey/StarGAN}} and Fader Networks\footnote{Fader Networks: \url{https://github.com/facebookresearch/Fader Networks}} are trained by their official code, while Shen et al. and CycleGAN are implemented by~ourself.

\subsection{Visual Analysis}
\textbf{Single Facial Attribute Editing.}
Firstly, we compare the proposed AttGAN with VAE/GAN~\cite{vaegan} and IcGAN~\cite{icgan} in terms of single facial attribute editing, shown in Fig.~\ref{fig:compare}.
As can be seen, in some cases VAE/GAN produces unexpected changes of other attributes, for example, all three male inputs become female in VAE/GAN when editing the blond hair attribute.
This phenomenon happens because the attribute vectors used for editing in VAE/GAN contains highly correlated attributes such as blond hair and female.
Therefore, some other unexpected but highly correlated attributes are also involved when using such attribute vectors for editing.
IcGAN performs better on accurately editing attributes, however, it seriously changes other attribute-excluding details especially the face identity.
This is mainly because IcGAN imposes attribute-independent constraint and normal distribution constraint on the latent representation, which harms its representation ability and results in loss of attribute-excluding information.
Compared to VAE/GAN and IcGAN, our AttGAN accurately edits both local attributes (bangs, eyeglasses and mouth open) and global attributes (gender), credited to the attribute classification constraint which guarantees the correct change of the attributes.
Moreover, AttGAN well preserves the attribute-excluding details such as face identity, illumination, and background, credited to that 1) the latent representation is constraint free, which guarantees its representation ability for conserving the attribute-excluding information, 2) the reconstruction learning explicitly enable the encoder-decoder to preserve the attribute-excluding details on the generated images.

Comparisons with StarGAN~\cite{stargan} are shown in Fig.~\ref{fig:compare_stargan}. As we can see, both StarGAN and AttGAN accurately edit attributes, but the StarGAN results contain some artifacts while the results of our AttGAN look more natural and realistic.

Comparisons with Fader Networks~\cite{fadernetworks}, Shen et al.~\cite{shen2016learning} and CycleGAN~\cite{cyclegan} are shown in Fig.~\ref{fig:compare_3}.
The results of Fader Networks especially on adding eyeglasses are blurry, which is very likely caused by the strict attribute-independent constraint on the latent representation.
The results of Shen et al. and CycleGAN contain noise and artifacts.
Another observation is that, adding ``Mustache'' makes the female (the second and fourth input in Fig.~\ref{fig:compare_3}) become male in Shen et al. and CycleGAN.
In the opposite, our AttGAN naturally add the mustache keeping the female's characteristic well although the model rarely (or never) sees the female with mustache in the training set, which reflects the AttGAN's superior ability to disentangle attributes (such as male and mustache) and preserve details.
%

\textbf{Multiple Facial Attribute Editing.}
All of VAE/GAN~\cite{vaegan}, IcGAN~\cite{icgan} and our AttGAN can simultaneously edit multiple attributes, and thus we investigate these three methods in terms of multiple facial attribute editing for more comprehensive comparison.
Fig.~\ref{fig:compare_multi} shows the results of simultaneously editing two or three attributes.

Similar to single attribute editing, some generated images from VAE/GAN contain undesired changes of other attributes since VAE/GAN cannot decorrelate highly correlated attributes.
As for IcGAN, distortion of face details and over smoothing become even more severe, because its constrained latent representation lead to worse performance in the more complex multiple attribute editing task.
By contrast, our method still performs well under complex combinations of attributes, benefited from the appropriate modeling of the relation between the attributes and the latent representation.


\begin{figure*}[t]
    \centering
    \includegraphics[width=1\linewidth]{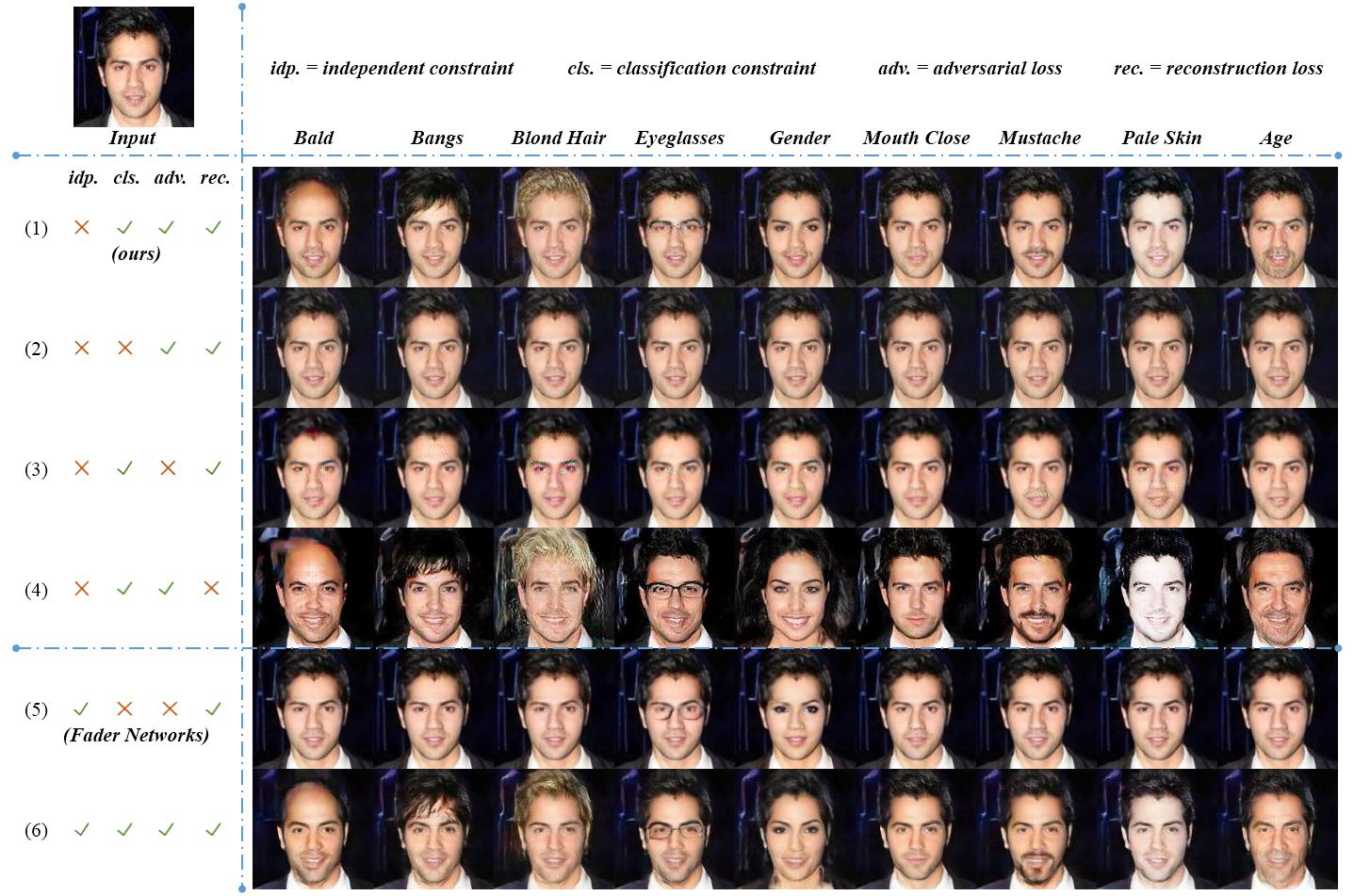}
    \caption{Effect of different combinations of the four components.}
    \label{fig:ablation_study}
\end{figure*}

\begin{figure*}[t]
    \centering
    \subfloat[Season Translation]{
        \centering
        \includegraphics[width=0.475\linewidth]{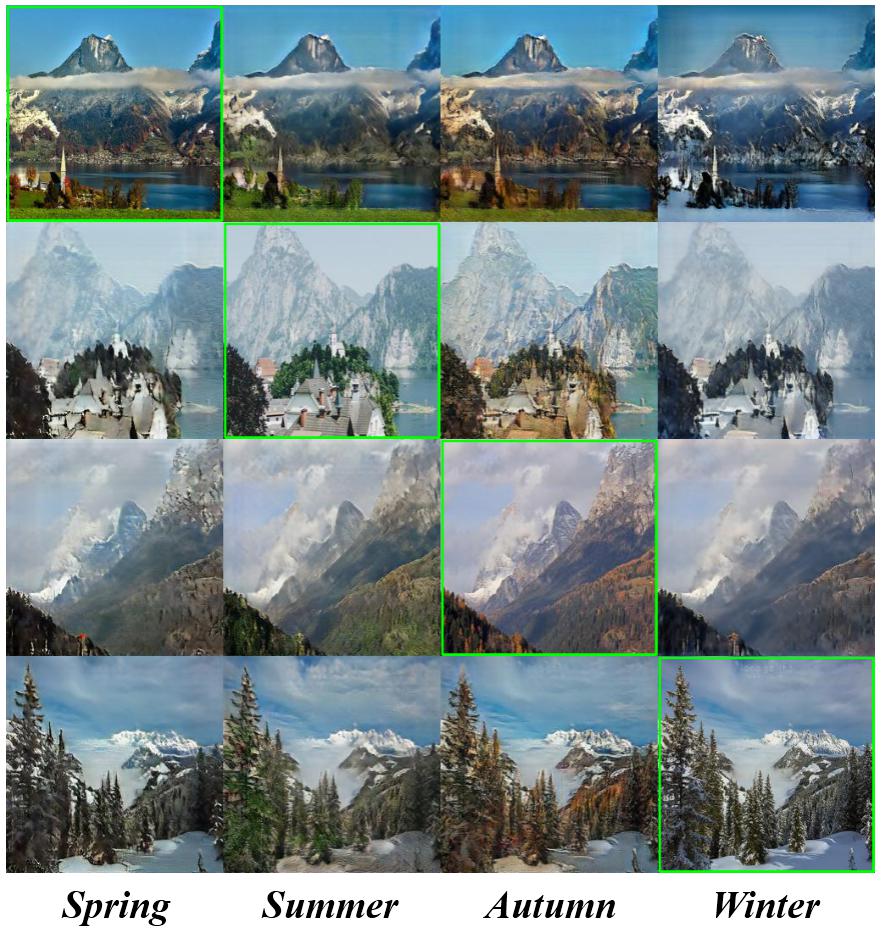}
        \label{fig:season}
    }
    \hfill
    \subfloat[Painting Translation]{
        \centering
        \includegraphics[width=0.475\linewidth]{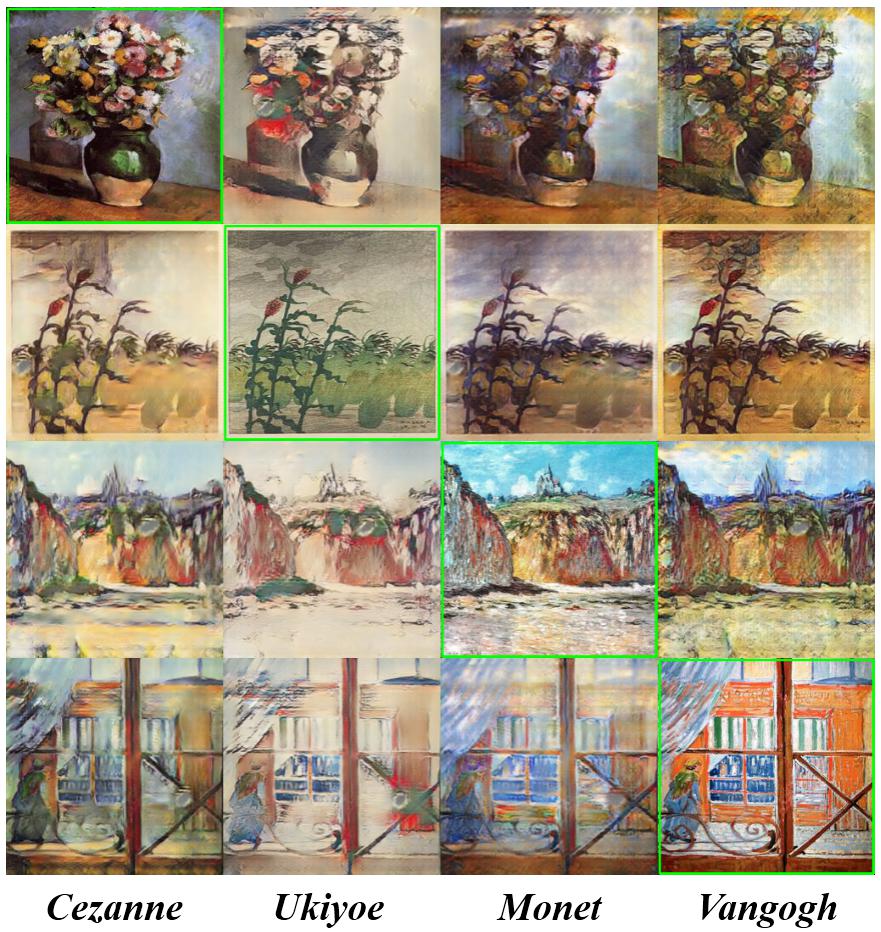}
        \label{fig:painting}
    }
    \caption{Exploration of AttGAN on image style translation. The \textbf{diagonal} ones are the inputs.}
    \label{fig:style_translation}
\end{figure*}

\vfill
\textbf{Attribute Intensity Control.}
Directly applicable for attribute intensity control is a characteristic of our AttGAN.
Although AttGAN is trained with binary attribute values (0/1), we find that AttGAN can be generalized for continuous attribute value in testing phase without any modification to its original design.
As shown in Fig.~\ref{fig:slide}, with continuous value in $[0, 1]$ as input, the gradual change of the generated images are smooth and natural.

\newpage
\textbf{Attribute Style manipulation.}
Fig.~\ref{fig:style} shows the results of the AttGAN extension for attribute style manipulation.
As can be seen, different styles of attributes are dug out, such as different sides of bangs: left, right or middle.
The extension is quite flexible and allows one to select the style he/she is interested in, rather than a stiff one.

\textbf{High Quality Results and Failures.}
Fig.~\ref{fig:384_1}-\ref{fig:384_3} in supplemental material shows additional results of high quality images with $384\times384$ resolution.
Fig.~\ref{fig:failures} in supplemental material shows some failures. These failures are often cased by the need of large appearance modification, such as editing a face with plenty of hair to ``Bald''.

\newpage
\subsection{Quantitative Analysis}
\textbf{Facial Attribute Editing Accuracy/Error.}
To evaluate the facial attribute editing accuracy of our AttGAN, an attribute classifier  independent of all methods is used to judge the attributes of the generated faces.
This attribute classifier is trained on CelebA~\cite{celeba} dataset and achieves average accuracy of 90.89\% per attribute on CelebA testing set.
If the attribute of a generated image is predicted the same as the desired one by the classifier, it is considered a correct generation, otherwise an incorrect one.
Besides, we also evaluate the average preservation error of the other attributes when editing each single attribute.

Fig.~\ref{fig:attribute_accuracy_1} shows the attribute editing accuracy of StarGAN~\cite{stargan}, VAE/GAN~\cite{vaegan}, IcGAN~\cite{icgan} and our AttGAN, all of which employ single model for multiple attribute editing.
%
As can be seen, both AttGAN and StarGAN achieve much better accuracy than VAE/GAN and IcGAN, especially on ``No Beard'', ``Pale Skin'' and ``Age''.
Moreover, the preservation errors of the other attributes of AttGAN and StarGAN are much lower than VAE/GAN and IcGAN as shown in Fig.~\ref{fig:attribute_error_1}.
%
As for the comparisons between AttGAN and StarGAN, the attribute editing accuracies of them are comparable, but the attribute preservation error of AttGAN is a bit higher.
However, the generated images of our AttGAN are much more natural and realistic than StarGAN (see Fig.~\ref{fig:compare_stargan})

Furthermore, Fig.~\ref{fig:attribute_accuracy_2} and Fig.~\ref{fig:attribute_error_2} show the attribute editing accuracy and preservation error of Fader Networks~\cite{fadernetworks}, Shen et al.~\cite{shen2016learning} and CycleGAN~\cite{cyclegan}, which employ one specific model for each attribute.
As can be seen, all three baselines well edit the attributes which is comparable to AttGAN, but their preservation errors of the other attributes are higher~than~AttGAN.


\subsection{Ablation Study: Effect of Each Component}

In this part, we evaluate the necessity of the three main components: attribute classification constraint, reconstruction loss and adversarial loss.
Besides, we also evaluate the disadvantage of the attribute-independent constraint.
In Fig.~\ref{fig:ablation_study}, we show the results of different combinations of these components, where all experiments are based on models which learn to handle multiple attributes with one network.
Row (1) contains the results of our AttGAN's original setting, which are natural and well preserve the attribute-excluding details.

Without the attribute classification constraint (row (2) of Fig.~\ref{fig:ablation_study}), the network just outputs the reconstruction images since these is no signal to force the network to generate the correct attributes.
Similar phenomenon (but with some noise) happens when we remove the adversarial loss although the classification constraint is kept (row (3)).
One possible reason is that the training with classification constraint but without adversarial loss is similar to making an adversarial attack~\cite{szegedy2013intriguing}.
Therefore, although the classification constraint exists, the adversarial examples with incorrect attributes still fool the classifier (by the noise).
In conclusion, the classification constraint does not work without the adversarial learning, or in other words, the adversarial learning helps to avoid adversarial examples.
However, this is another topic needing more theoretical analysis and experiments, which is far beyond this paper.

In row (4) of Fig.~\ref{fig:ablation_study}, we present the results of AttGAN without reconstruction loss. As shown, although the resulting attributes are correct, the face identities change a lot accompanied with many artifacts. Therefore, the reconstruction loss is vital for preserving the attribute-excluding details.

Row (5) of Fig.~\ref{fig:ablation_study} presents the results of the Fader Networks~\cite{fadernetworks} like setting (attribute-independent constraint + reconstruction learning) and row (6) is AttGAN with attribute-independent constraint.
As we can see in the row (5), the Fader Networks like setting works only on eyeglasses, gender and mouth open attributes with unsatisfactory performance. When we combine the AttGAN losses with the Fader Networks losses (row (6)), the attributes is correctly edited but the results contain artifacts and the attribute-excluding details change (e.g., the shape of nose and mouth).
These experiments demonstrates that the attribute-independent constraint on the latent representation is not a favorable solution for facial attribute editing, since it constraints the representation ability of the latent code resulting in information loss and degraded output images.

\subsection{Exploration of Image Translation}
Since facial attribute editing is closely related to image translation, we also try our AttGAN on the image style translation task where we define the style as a kind of attribute.
We employ AttGAN on a season dataset~\cite{combogan} and a painting dataset~\cite{cyclegan} and the results are shown in Fig.~\ref{fig:style_translation}.
As we can see, the results of season are acceptable but the style translation of paintings is not so good accompanied with artifacts and blurriness.
%
Compared to facial attribute editing, image style translation needs more variations on texture and color, a single model might be difficult to simultaneously handle all styles with large variation.
However, AttGAN is a potential framework which deserves more explorations~and~extensions.

\vfill
\section{Conclusion}
From the perspective of facial attribute editing, we reveal and validate the disadvantage of the attribute-independent constraint on the latent representation.
Further, we properly consider the relation between the attributes and the latent representation and propose an AttGAN method, which incorporates the attribute classification constraint, the reconstruction learning, and the adversarial learning to form an effective framework for high quality facial attribute editing.
Experiments demonstrate that our AttGAN can accurately edit facial attributes, while well preserving the attribute-excluding details, with better visual effect, editing accuracy and lower editing error than the competing methods.
Moreover, our AttGAN is directly applicable for attribute intensity control and can be extended for attribute style manipulation, which shows its potential for further exploration.


%



\section*{Acknowledgment}
This work was supported partly by National Key R\&D Program of China under contract No.2017YFA0700800, Natural Science Foundation of China under contracts Nos.61390511, 61650202, and 61402443.


\ifCLASSOPTIONcaptionsoff
  \newpage
\fi



\bibliographystyle{IEEEtran}
\bibliography{reference}
\vfill
\begin{IEEEbiography}[{\includegraphics[width=1in,height=1.25in,clip,keepaspectratio]{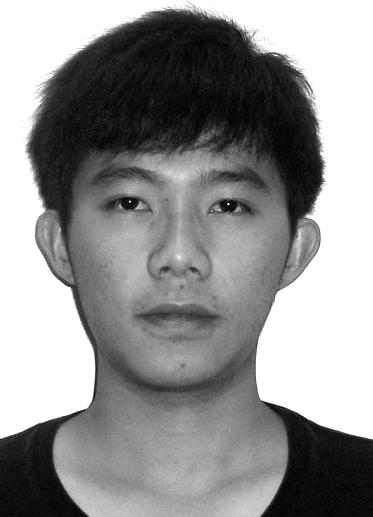}}]{Zhenliang He}
received the B.E. degree from Beijing University of Posts and Telecommunications and is pursing the Ph.D. degree from Institute of Computing Technology (ICT), Chinese Academy of Sciences (CAS), Beijing, China. His research interests include pattern recognition, machine learning and computer vision.
\end{IEEEbiography}

\vfill
\begin{IEEEbiography}[{\includegraphics[width=1in,height=1.25in,clip,keepaspectratio]{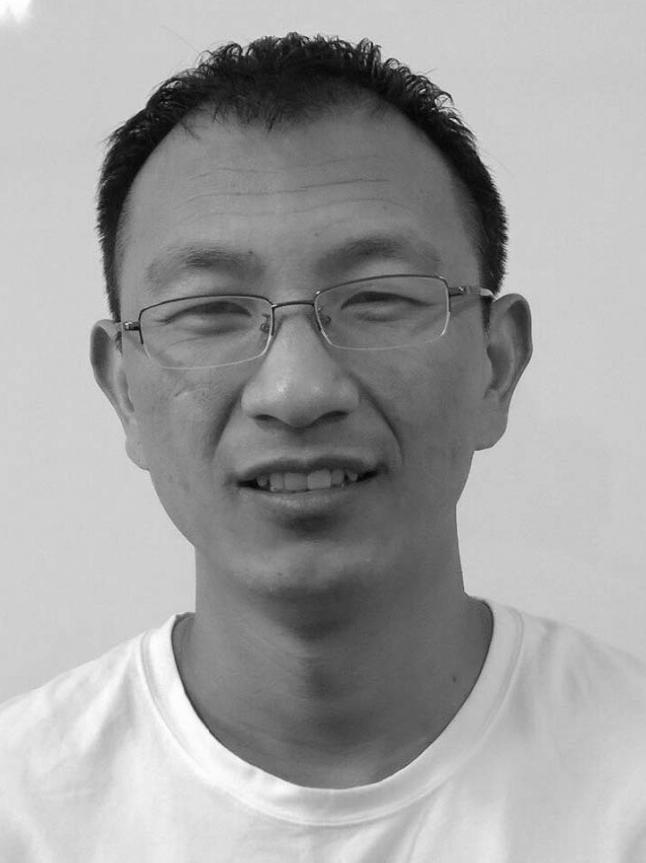}}]{Wangmeng Zuo}(M'09-SM'14)
received the Ph.D. degree in computer application technology from the Harbin Institute of Technology, Harbin, China, in 2007. He is currently a Professor in the School of Computer Science and Technology, Harbin Institute of Technology. His current research interests include image enhancement and restoration, image and face editing, object detection, visual tracking, and image classification. He has published over 70 papers in toptier academic journals and conferences. He has served as a Tutorial Organizer in ECCV 2016, an Associate Editor of the \emph{IET Biometrics and Journal of Electronic Imaging}.
\end{IEEEbiography}

\vfill
\begin{IEEEbiography}[{\includegraphics[width=1in,height=1.25in,clip,keepaspectratio]{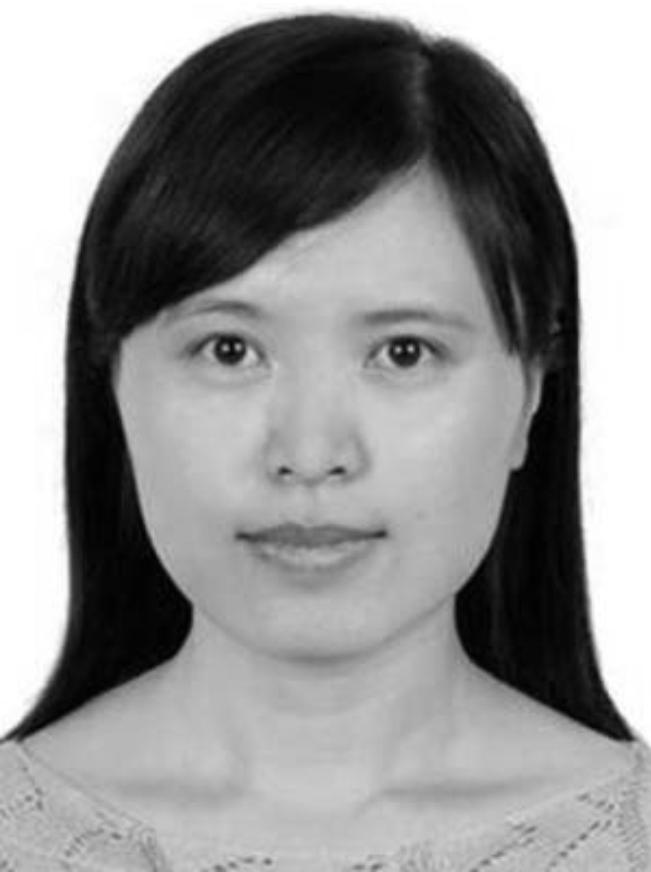}}]{Meina Kan}
is now an Associate Professor with the Institute of Computing Technology (ICT), Chinese Academy of Sciences (CAS), where she received the Ph.D. degree in computer science in 2013. Her research mainly focuses on face detection, face recognition, transfer learning and deep learning.
\end{IEEEbiography}

\vfill
\begin{IEEEbiography}[{\includegraphics[width=1in,height=1.25in,clip,keepaspectratio]{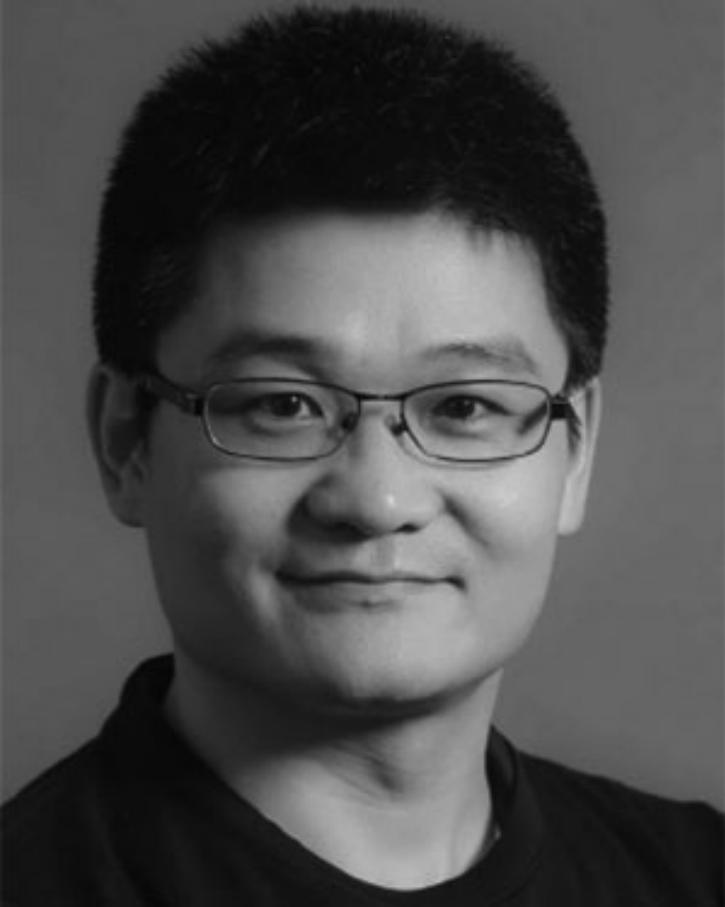}}]{Shiguang Shan}
is a professor of ICT, CAS, and the deputy director with the Key Laboratory of Intelligent Information Processing, CAS. His research interests cover computer vision, pattern recognition, and machine learning. He has authored more than 200 papers in refereed journals and proceedings in the areas of computer vision and pattern recognition. He was a recipient of the China's State Natural Science Award in 2015, and the China's State S\&T Progress Award in 2005 for his research work. He has served as the Area Chair for many international conferences, including ICCV'11, ICPR'12, ACCV'12, FG'13, ICPR'14, and ACCV'16. He is an associate editor of several journals, including the \emph{IEEE Transactions on Image Processing}, the \emph{Computer Vision and Image Understanding}, the \emph{Neurocomputing}, and the \emph{Pattern Recognition Letters}. He is a senior member of the IEEE.
\end{IEEEbiography}

\vfill
\begin{IEEEbiography}[{\includegraphics[width=1in,height=1.25in,clip,keepaspectratio]{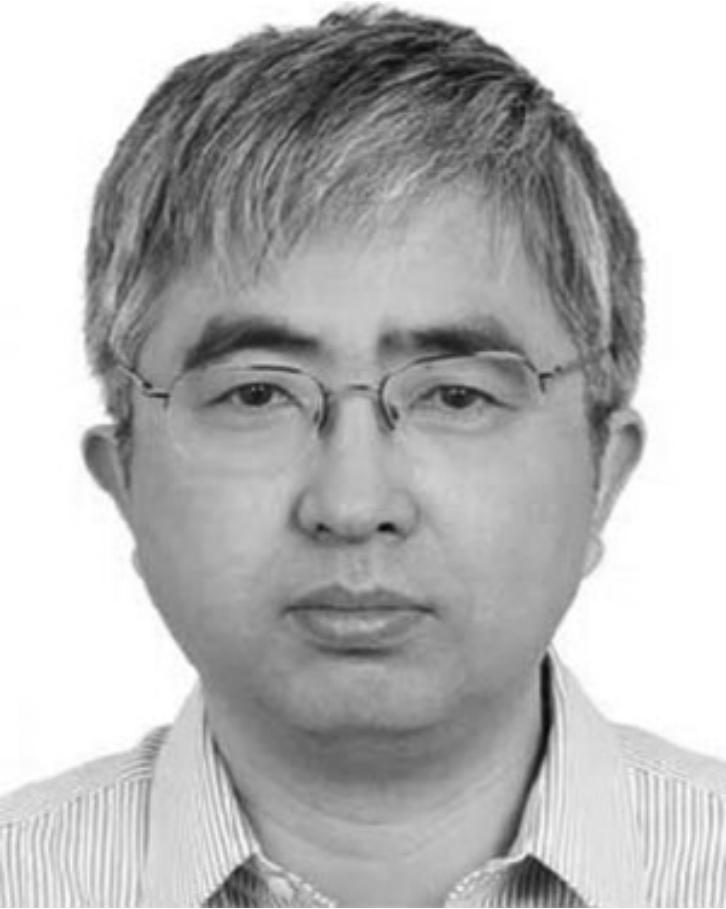}}]{Xilin Chen}
is a professor of ICT, CAS. He has authored one book and more than 200 papers in refereed journals and proceedings in the areas of computer vision, pattern recognition, image processing, and multimodal interfaces. He served as an Organizing Committee/Program Committee member for more than 70 conferences. He was a recipient of several awards, including the China's State Natural Science Award in 2015, the China's State S\&T Progress Award in 2000, 2003, 2005, and 2012 for his research work. He is currently an associate editor of the \emph{IEEE Transactions on Multimedia}, a leading editor of the \emph{Journal of Computer Science and Technology}, and an associate editor-in-chief of the \emph{Chinese Journal of Computers}. He is a fellow of the China Computer Federation (CCF), IAPR, and the IEEE.
\end{IEEEbiography}

\clearpage
\markboth{SUPPLEMENTAL MATERIAL}{}
\begin{figure*}[p]
    \parbox[c][0.99\textheight]{1\linewidth}{
        \centering
        \subfloat[Add Bangs]{
            \includegraphics[width=0.9\linewidth]{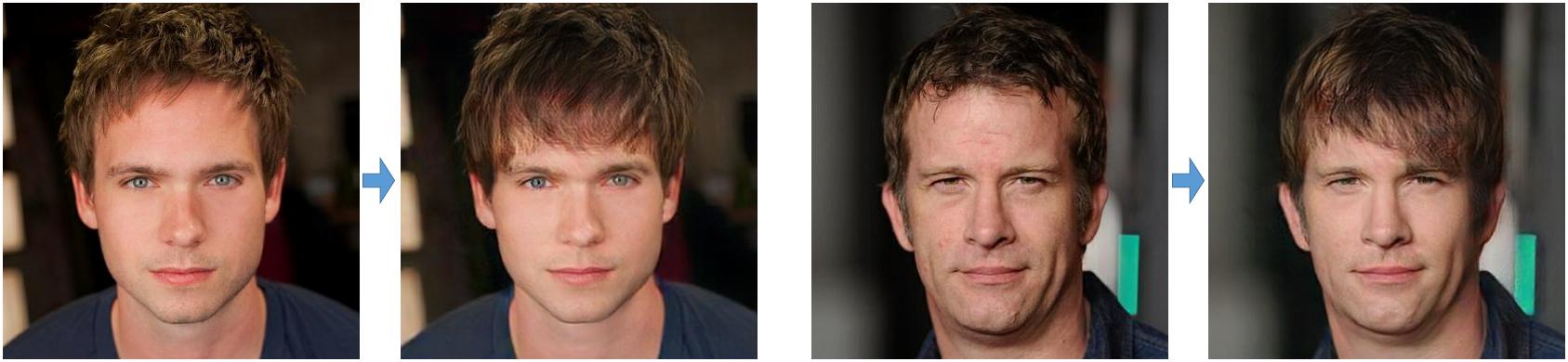}
        }
        \vfill
        \vspace{-7pt}
        \subfloat[Remove Bangs]{
            \includegraphics[width=0.9\linewidth]{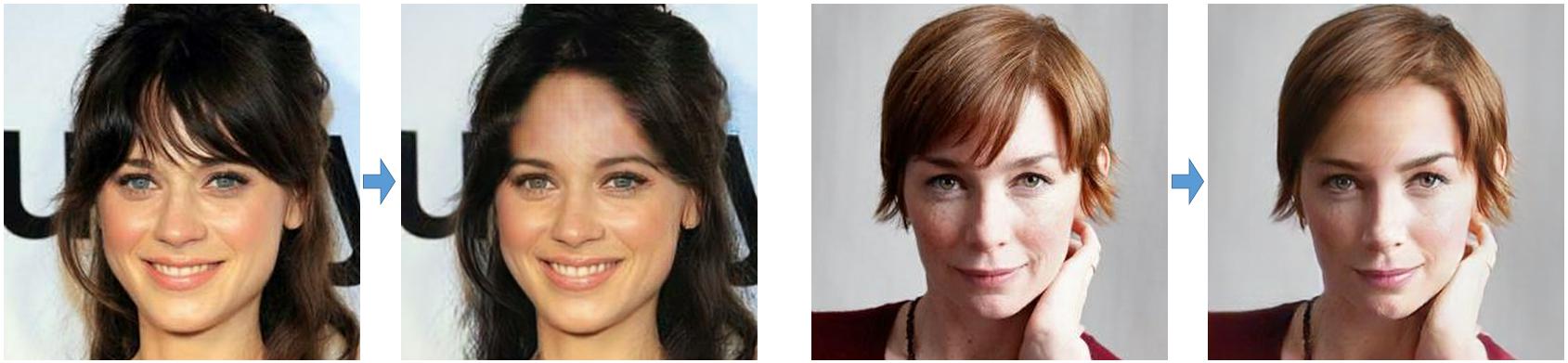}
        }
        \vfill
        \vspace{-7pt}
        \subfloat[Add Eyeglasses]{
            \includegraphics[width=0.9\linewidth]{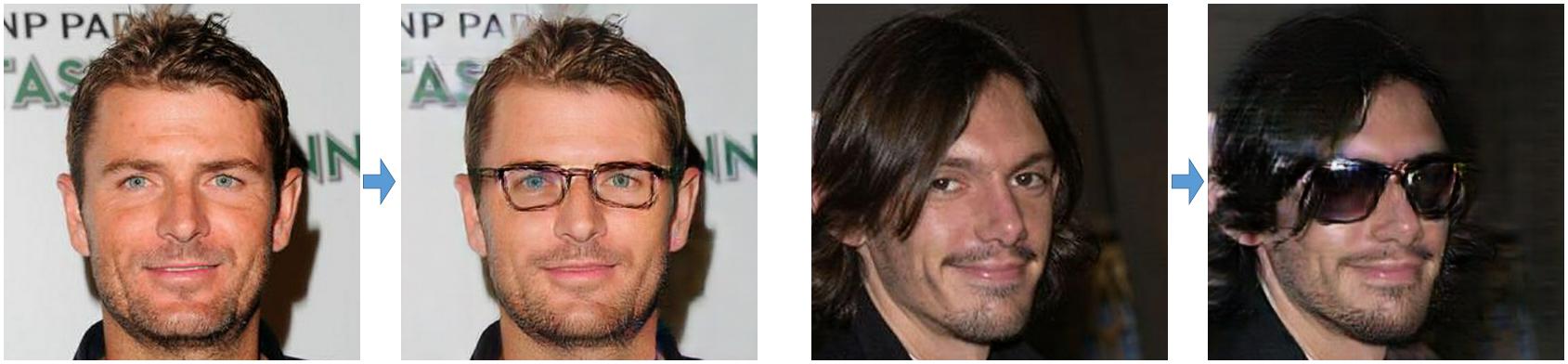}
        }
        \vfill
        \vspace{-7pt}
        \subfloat[Remove Eyeglasses]{
            \includegraphics[width=0.9\linewidth]{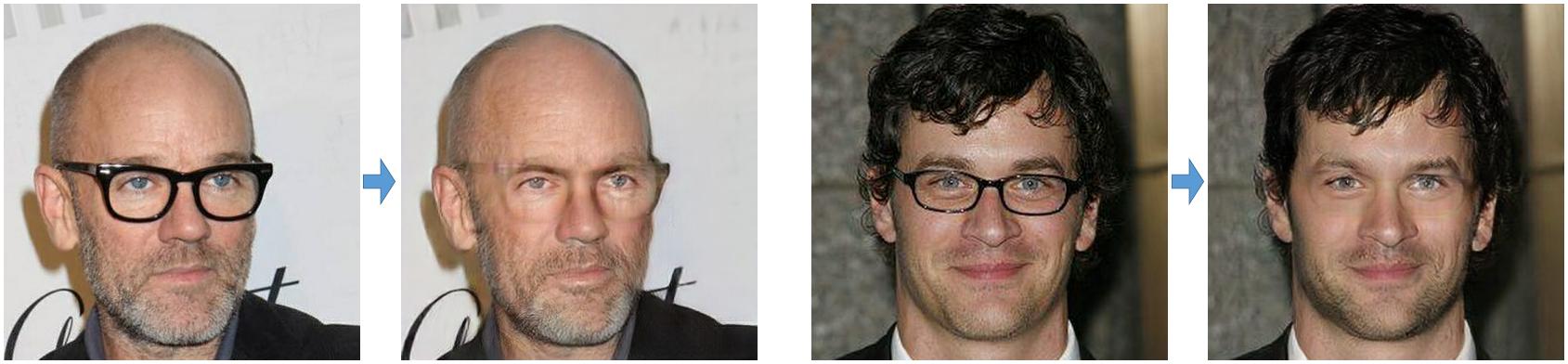}
        }
        \vfill
        \vspace{-7pt}
        \subfloat[Add Beard]{
            \includegraphics[width=0.9\linewidth]{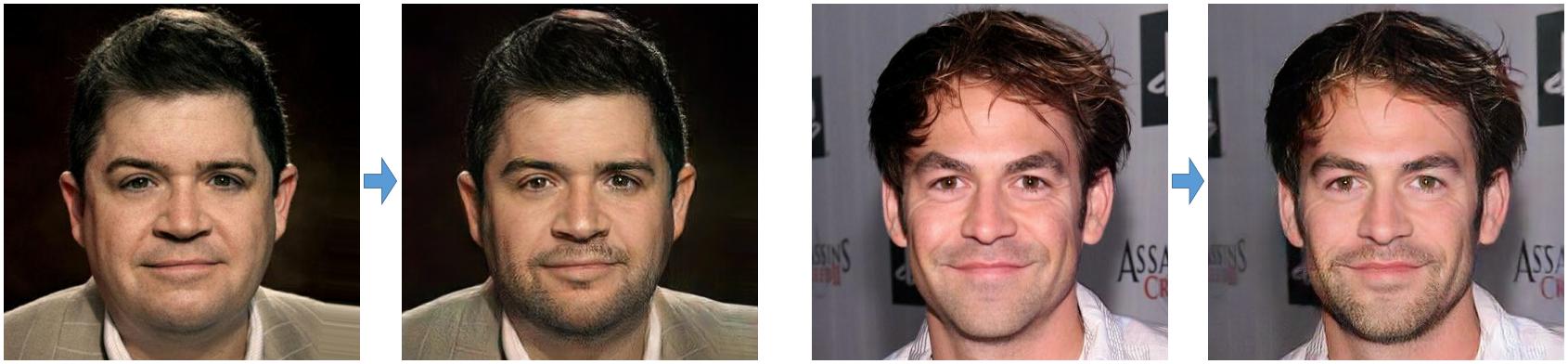}
        }
        \vfill
        \vspace{-7pt}
        \subfloat[Remove Beard]{
            \includegraphics[width=0.9\linewidth]{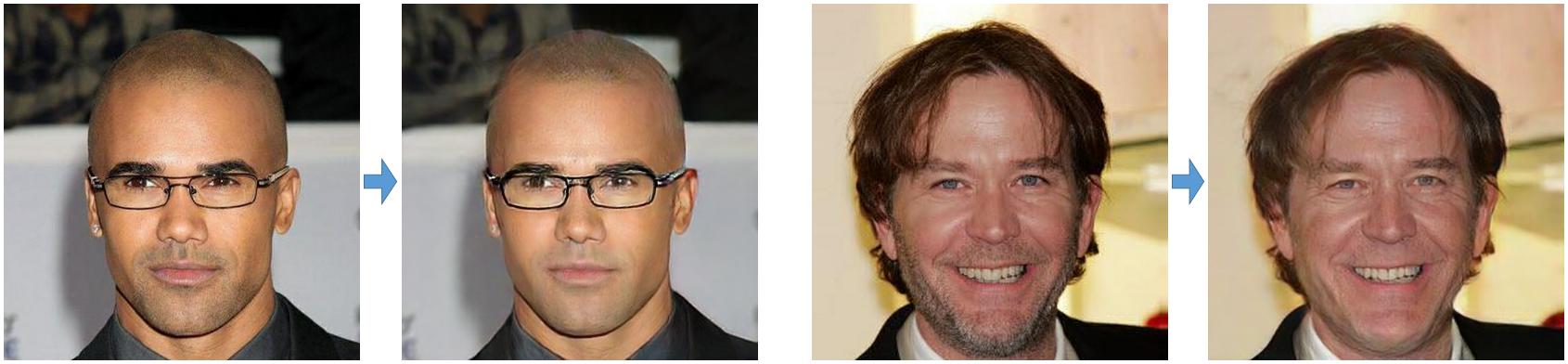}
        }
        \vspace{-7pt}
        \caption{Additional AttGAN results of high quality images with $384\times384$ resolution. Zoom in for better resolution.}
        \label{fig:384_1}
    }
\end{figure*}

\clearpage
\markboth{}{}
\begin{figure*}[p]
    \parbox[c][0.99\textheight]{1\linewidth}{
        \centering
        \subfloat[To Female]{
            \includegraphics[width=0.9\linewidth]{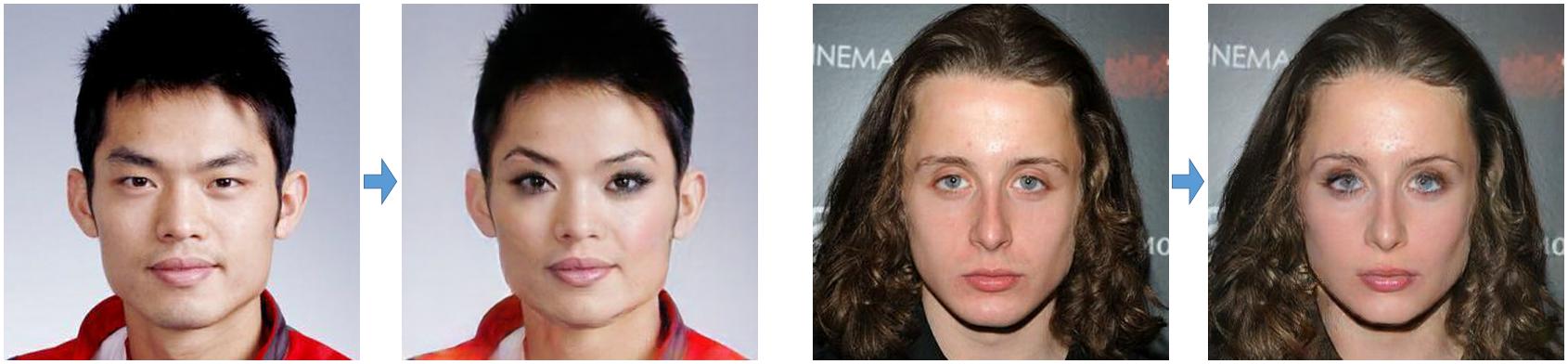}
        }
        \vfill
        \vspace{-3pt}
        \subfloat[To Male]{
            \includegraphics[width=0.9\linewidth]{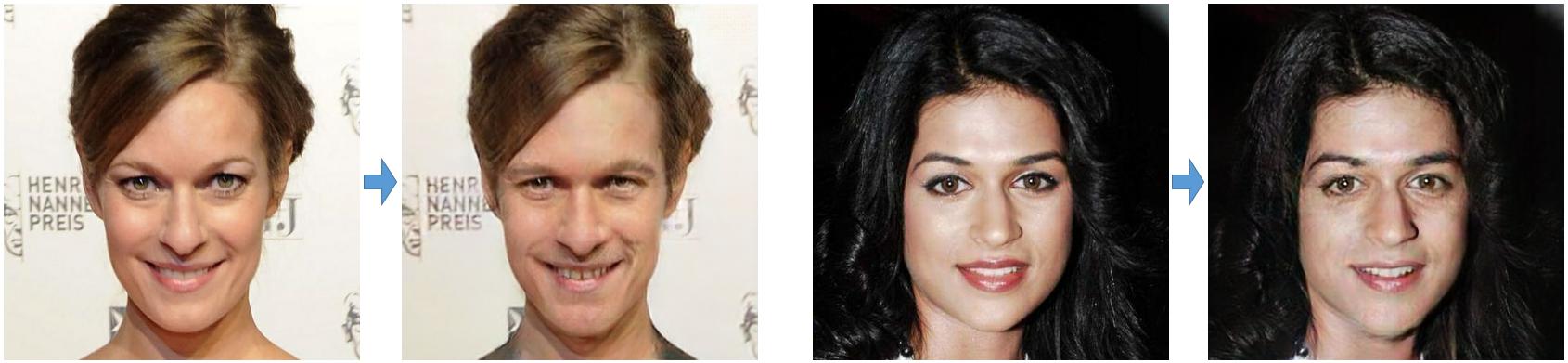}
        }
        \vfill
        \vspace{-3pt}
        \subfloat[To Black Hair]{
            \includegraphics[width=0.9\linewidth]{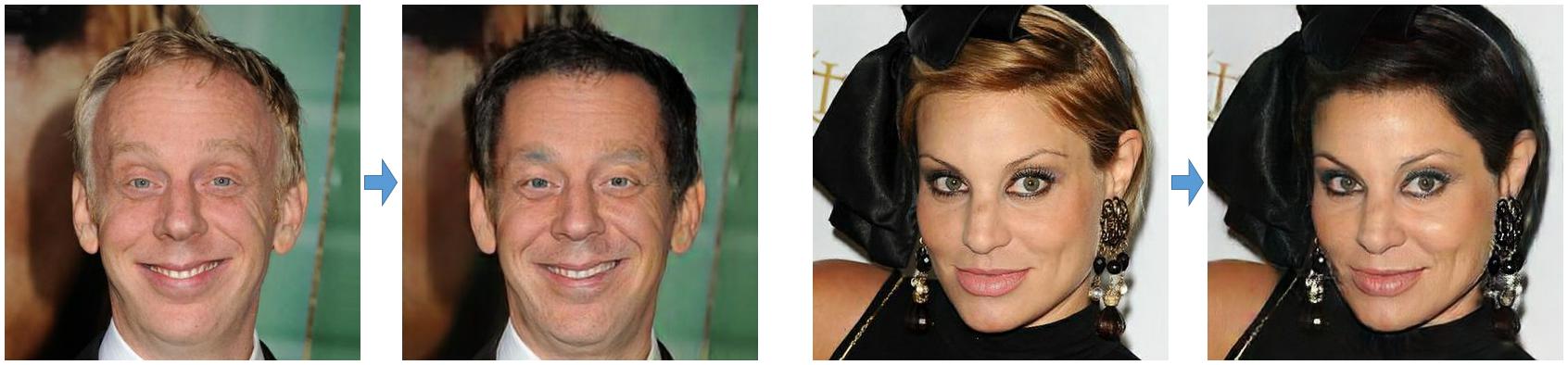}
        }
        \vfill
        \vspace{-3pt}
        \subfloat[To Blond Hair]{
            \includegraphics[width=0.9\linewidth]{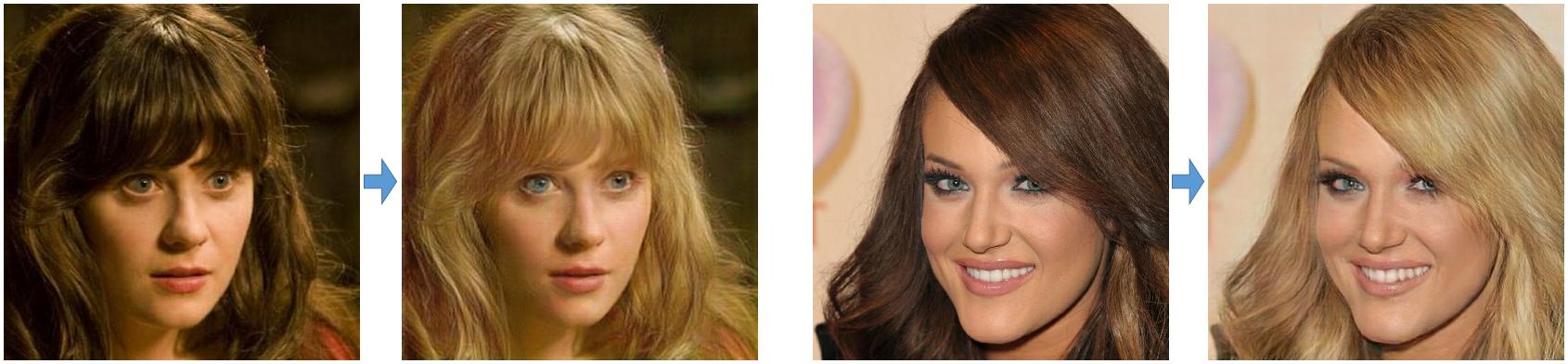}
        }
        \vfill
        \vspace{-3pt}
        \subfloat[\scriptsize To Bushy Eyebrows + Mouth Open]{
            \includegraphics[width=0.44\linewidth]{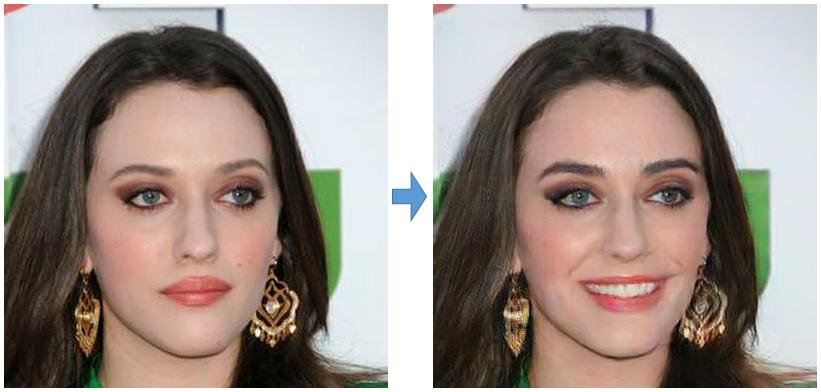}
        }
        \vspace{-3pt}
        \subfloat[\scriptsize To Bushy Eyebrows + Mouth Close]{
            \includegraphics[width=0.44\linewidth]{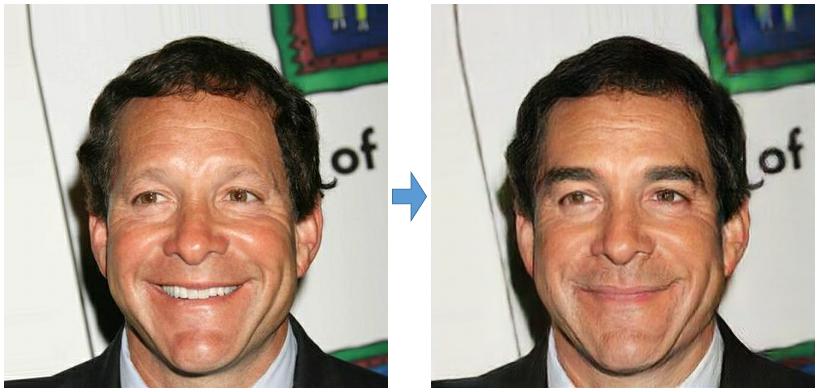}
        }
        \vfill
        \vspace{-3pt}
        \subfloat[\scriptsize To Light Eyebrows + Mouth Open]{
            \includegraphics[width=0.44\linewidth]{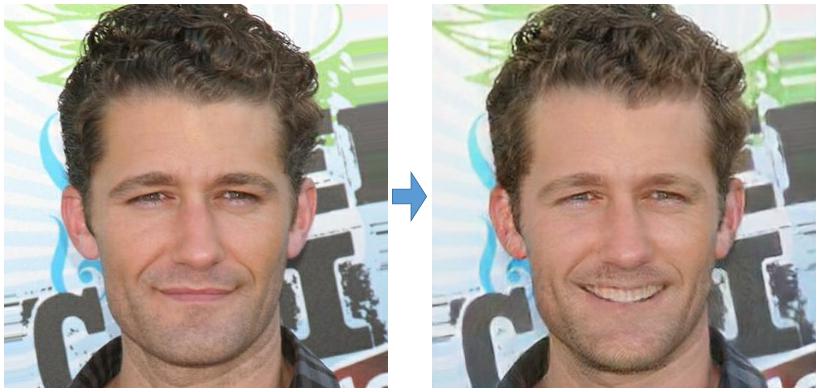}
        }
        \vspace{-3pt}
        \subfloat[\scriptsize To Light Eyebrows + Mouth Close]{
            \includegraphics[width=0.44\linewidth]{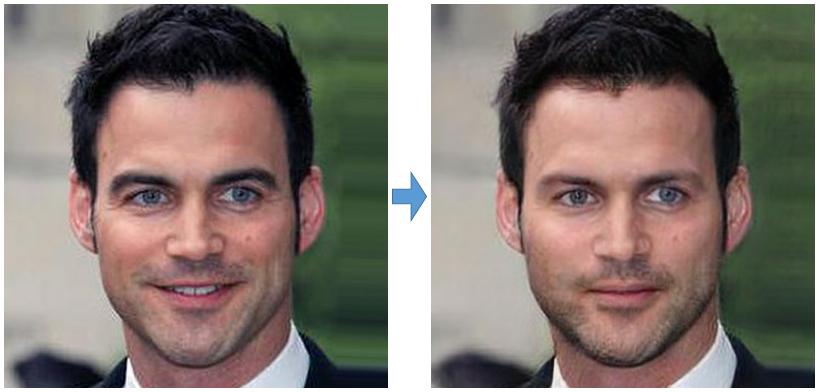}
        }
        \vspace{-2pt}
        \caption{Additional AttGAN results of high quality images with $384\times384$ resolution. Zoom in for better resolution.}
        \label{fig:384_2}
    }
\end{figure*}

\begin{figure*}[p]
    \parbox[c][0.99\textheight]{1\linewidth}{
        \centering
        \subfloat[\scriptsize To Male + To Young]{
            \includegraphics[width=0.44\linewidth]{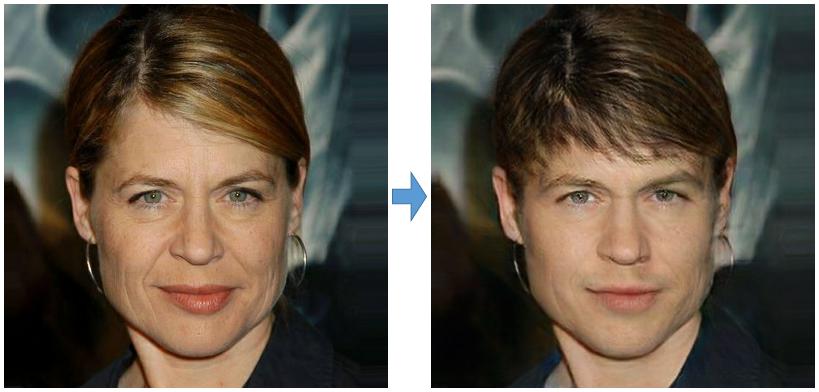}
        }
        \subfloat[\scriptsize To Male + To Old]{
            \includegraphics[width=0.44\linewidth]{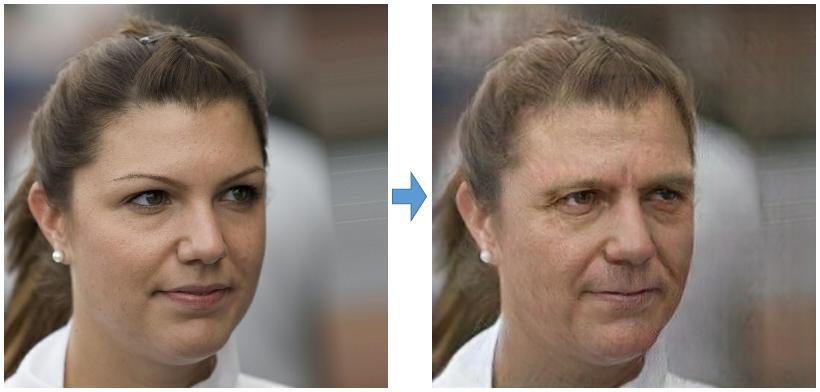}
        }
        \vfill
        \vspace{-3pt}
        \subfloat[\scriptsize To Female + To Young]{
            \includegraphics[width=0.44\linewidth]{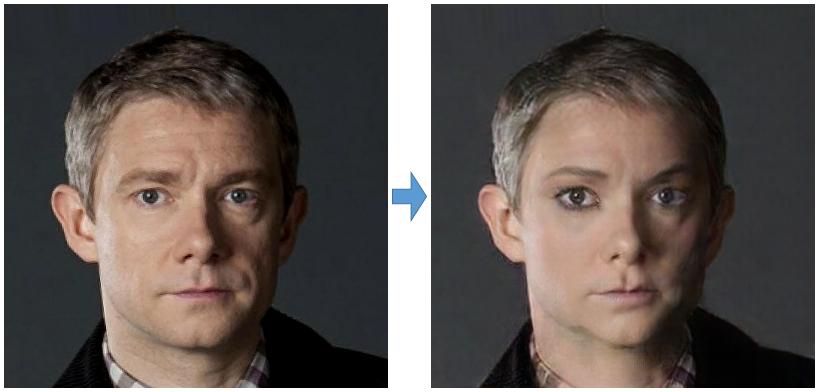}
        }
        \vspace{-3pt}
        \subfloat[\scriptsize To Female + To Old]{
            \includegraphics[width=0.44\linewidth]{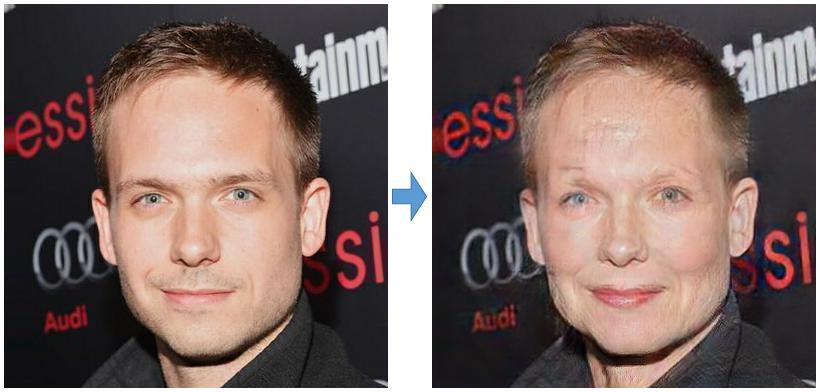}
        }
        \vfill
        \vspace{-3pt}
        \subfloat[\scriptsize To Blond Hair + Add Beard]{
            \includegraphics[width=0.44\linewidth]{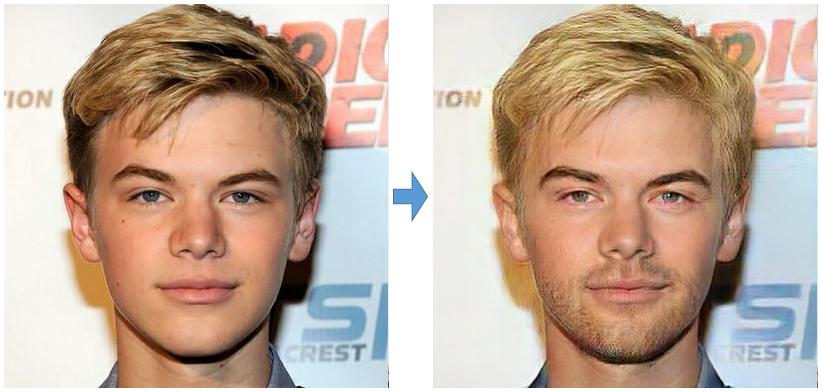}
        }
        \vspace{-3pt}
        \subfloat[\scriptsize To Blond Hair + Remove Beard]{
            \includegraphics[width=0.44\linewidth]{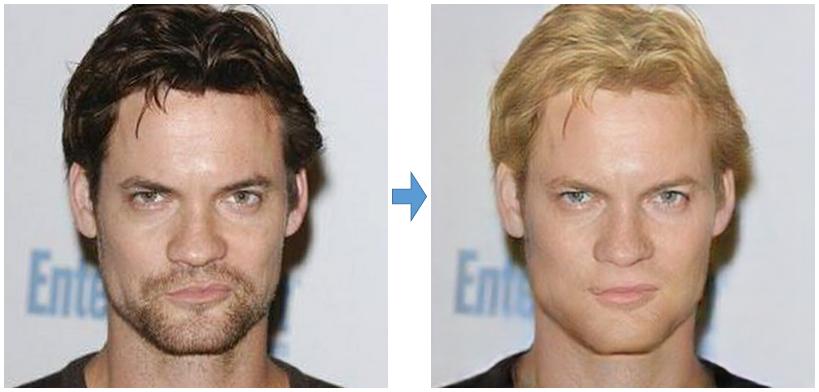}
        }
        \vfill
        \vspace{-3pt}
        \subfloat[\scriptsize To Brown Hair + Add Beard]{
            \includegraphics[width=0.44\linewidth]{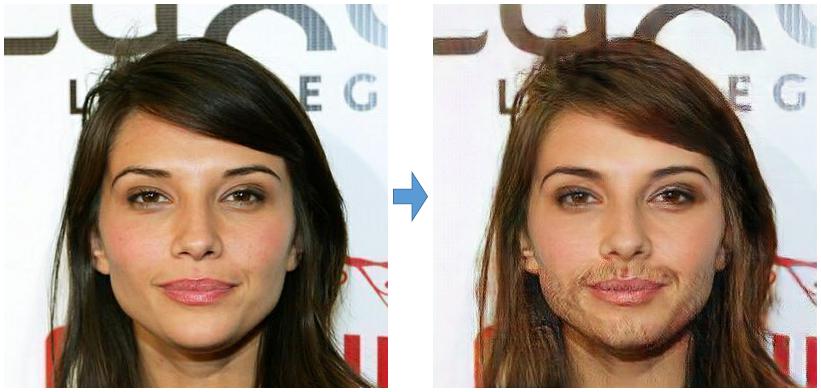}
        }
        \vspace{-3pt}
        \subfloat[\scriptsize To Brown Hair + Remove Beard]{
            \includegraphics[width=0.44\linewidth]{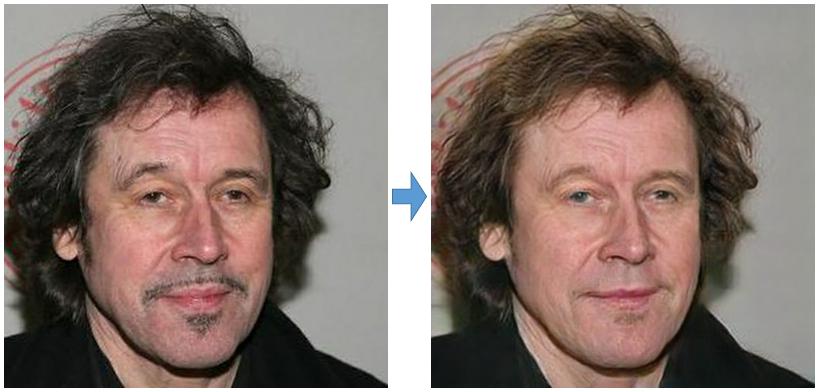}
        }
        \vspace{-4pt}
        \caption{Additional AttGAN results of high quality images with $384\times384$ resolution. Zoom in for better resolution.}
        \label{fig:384_3}
        \vfill
        \subfloat[To Bald]{
            \includegraphics[width=0.44\linewidth]{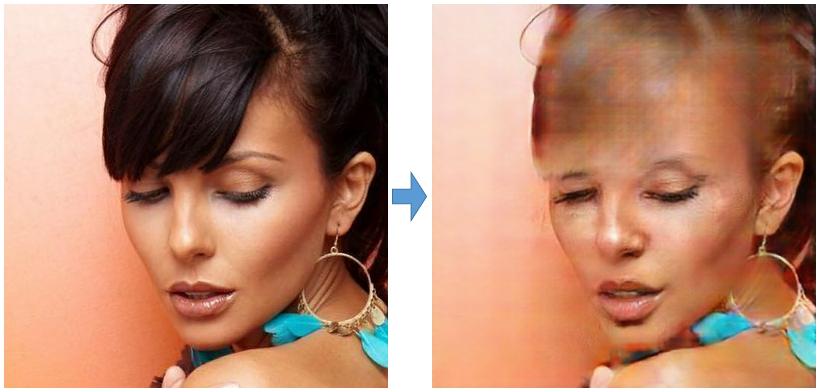}
        }
        \vspace{-3pt}
        \subfloat[Add Bangs]{
            \includegraphics[width=0.44\linewidth]{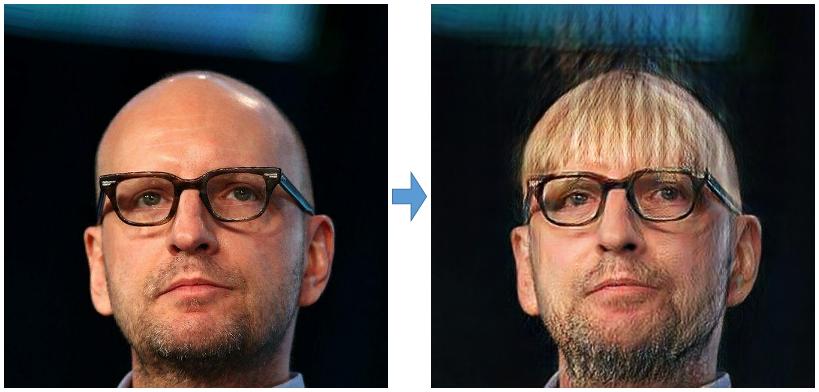}
        }
        \vfill
        \vspace{-3pt}
        \subfloat[To Black Hair]{
            \includegraphics[width=0.44\linewidth]{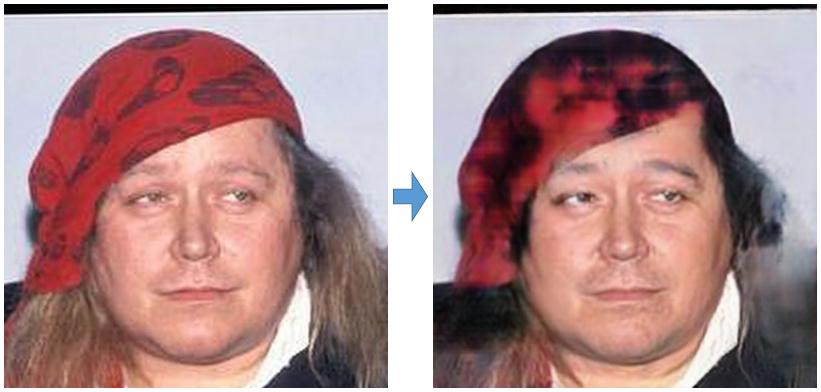}
        }
        \vspace{-3pt}
        \subfloat[Remove Eyeglasses]{
            \includegraphics[width=0.44\linewidth]{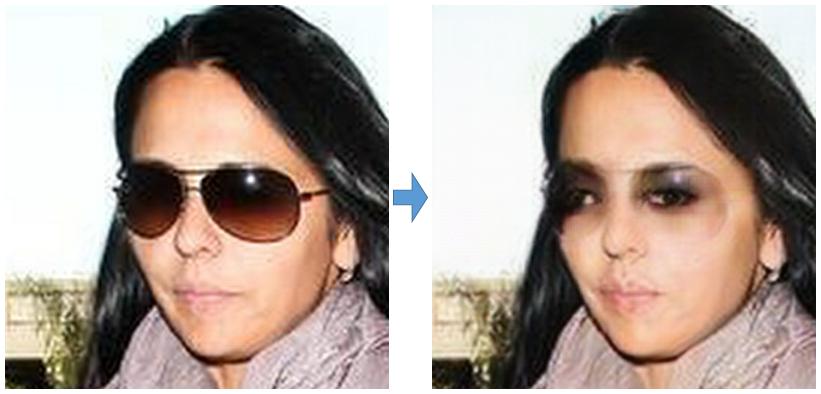}
        }
        \vspace{-4pt}
        \caption{Failures, which are often caused by the need of large appearance modification.}
        \label{fig:failures}
    }
\end{figure*}

\end{document}